\documentclass[10pt,twocolumn,letterpaper]{article}

\usepackage[pagenumbers]{cvpr} 

\usepackage{microtype}
\definecolor{cvprblue}{rgb}{0.21,0.49,0.74}
\usepackage[pagebackref,breaklinks,colorlinks,allcolors=cvprblue]{hyperref}
\usepackage[table,xcdraw]{xcolor}
\usepackage{booktabs,siunitx,adjustbox}
\usepackage{graphicx}
\usepackage{multirow}
\usepackage{subcaption}
\def\paperID{} 
\def\confName{CVPR}
\def\confYear{2026}

\title{BeautyGRPO: Aesthetic Alignment for Face Retouching via Dynamic Path Guidance and Fine-Grained Preference Modeling}

\author{
Jiachen Yang$^{1}$ \quad Xianhui Lin$^{2\dagger}$ \quad Yi Dong$^{2}$ \quad Zebiao Zheng$^{2}$ \\
Xing Liu$^{2}$ \quad Hong Gu$^{2}$ \quad Yanmei Fang$^{1,3*}$ \\[2mm]
$^{1}$School of Cyber Science and Technology, Shenzhen Campus of Sun Yat-sen University, China \\[1.2mm]
$^{2}$vivo BlueImage Lab, vivo Mobile Communication Co., Ltd., China \\[1.2mm]
$^{3}$Guangdong Provincial Key Laboratory of Information Security Technology, China \\[1mm]
\href{https://beautygrpo.github.io/}{\textbf{Project Page}} \\[1mm]
{\footnotesize $^\dagger$Project leader \quad $^*$Corresponding author}
}

\begin{document}
\maketitle
\begin{abstract}
Face retouching requires removing subtle imperfections while preserving unique facial identity  features, in order to enhance overall aesthetic appeal. However, existing methods suffer from a fundamental trade-off. Supervised learning on labeled data is constrained to pixel-level label mimicry, failing to capture complex subjective human aesthetic preferences. Conversely, while online reinforcement learning (RL) excels at preference alignment, its stochastic exploration paradigm conflicts with the high-fidelity demands of face retouching and often introduces noticeable noise artifacts due to accumulated stochastic drift.
To address these limitations, we propose BeautyGRPO, a reinforcement learning framework that aligns face retouching with human aesthetic preferences. We construct FRPref-10K, a fine-grained preference dataset covering five key retouching dimensions, and train a specialized reward model capable of evaluating subtle perceptual differences. To reconcile exploration and fidelity, we introduce Dynamic Path Guidance (DPG). DPG stabilizes the stochastic sampling trajectory by dynamically computing an anchor-based ODE path and replanning a guided trajectory at each sampling timestep, effectively correcting stochastic drift while maintaining controlled exploration.
Extensive experiments show that BeautyGRPO outperforms both specialized face retouching methods and general image editing models, achieving superior texture quality, more accurate blemish removal, and overall results that better align with human aesthetic preferences.
\end{abstract}    
\section{Introduction}
\label{sec:intro}

\begin{figure}[htbp] 
\begin{center}
\includegraphics[width=\linewidth]{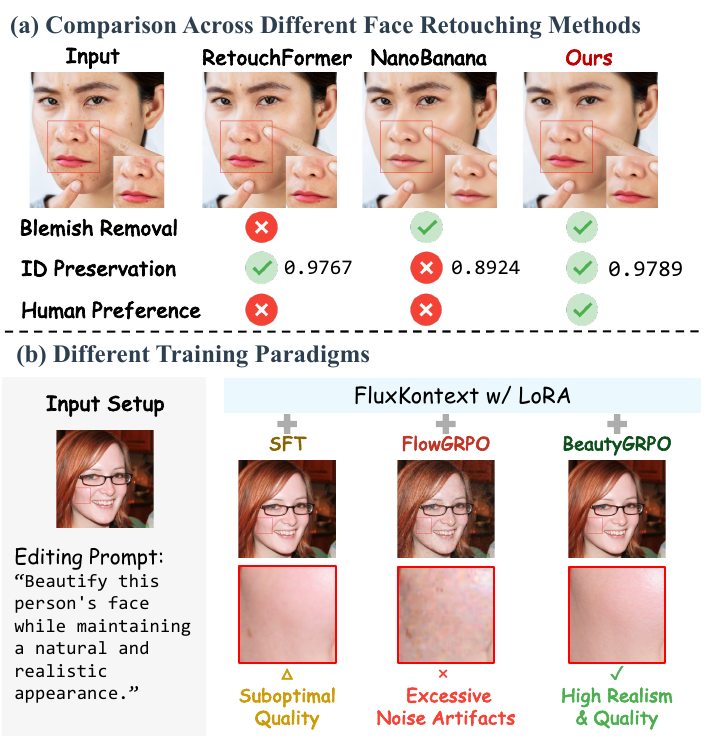}
\end{center}
\caption{Comparison of face retouching methods and training paradigms. (a) Our approach outperforms specialized (RetouchFormer) and general (NanoBanana) models in blemish removal, identity preservation, and human preference. (b) By stabilizing stochastic exploration, BeautyGRPO overcomes the limitations of Supervised Fine-Tuning (SFT) and standard online RL (FlowGRPO), achieving natural and highly realistic results.}
\label{fig1}
\end{figure}

With the rapid proliferation of social media and imaging devices, digital portraiture has become an integral part of daily life, driving a growing demand for high-quality face retouching. This task is inherently complex, as it requires removing subtle imperfections such as acne and blemishes while preserving identity features such as moles, pores, and natural skin texture, in order to enhance overall aesthetic appeal. In essence, face retouching must strike a balance between high visual fidelity to the original portrait and consistency with subjective human aesthetic preferences.

To address this challenge, researchers have explored a variety of specialized architectures, ranging from CNN-based \cite{shafaei2021autoretouch,lei2022abpn} to advanced Transformer-based models \cite{wang2022restoreformer, wang2023restoreformer++, xue2024vretoucher, wen2024retouchformer,xue2025retouchgpt}. 
Recently, general image editing models \cite{liu2025step1x,zhang2025ICEdit,openai2025gptimage,google2025nanobanana,seedream2025seedream} have also shown strong potential for face retouching. However, most existing methods still depend on supervised learning, either training from scratch or supervised fine-tuning (SFT). The limitation of such supervised objectives lies not only in the scarcity of curated data, but also in their pixel-level learning targets. They train models to imitate labeled reference images rather than to align with human aesthetic preferences. This objective enforces local fidelity while neglecting perceptual alignment, leading to results that may be faithful to the labels but still misaligned with human aesthetic expectations. As a result, the outputs often exhibit visually rigid or unnatural retouching effects, as illustrated in Fig.~\ref{fig1}. Moreover, by constraining optimization to pixel-level reconstruction, supervised learning encourages overfitting to specific retouching styles and prevents the model from discovering solutions that surpass the aesthetic quality of the training data. This observation raises a fundamental question: \textbf{Can we align face retouching with human preferences?}

Reinforcement learning (RL) has recently emerged as a promising framework for aligning image generation with human preferences \cite{prabhudesai2023aligning,clark2024directly,prabhudesai2024video,DDPO2024,fan2024dpok,rafailov2023direct,wallace2024diffusion,yang2024D3PO,yuan2024self,zhu2025dspo,guo2025can,Wang2025DiffusionNPO}. In text-to-image (T2I) generation, methods such as FlowGRPO \cite{liu2025flow} demonstrate strong controllability by integrating online RL into flow-matching models \cite{lipman2023flow,liu2023rectifiedflow,esser2024scaling,flux2023}. This online RL mechanism enables active exploration, allowing the model to discover solutions beyond those observed during supervised training and to better align with human judgment. Despite these successes, directly applying T2I-RL frameworks to face retouching introduces two fundamental challenges. The first is the conflict between fidelity and exploration. To encourage exploration, FlowGRPO injects stochastic noise into the sampling process, converting the deterministic ODE trajectory into a stochastic differential equation (SDE). While this stochasticity enhances diversity in T2I generation, it is detrimental to face retouching, where visual stability and precision are essential. As shown in Fig.~\ref{fig1}, the accumulated noise causes the trajectory to deviate from the high-fidelity manifold and produce noticeable noise artifacts. The second challenge lies in the granularity of reward supervision. Existing reward models for T2I generation or general image editing \cite{wu2023HPSv2,xu2023imagereward,xu2024visionreward,luo2025editscore,wu2025editreward,wang2025unified} mainly focus on global aesthetics or instruction following. They lack the fine-grained perceptual sensitivity required for face retouching, which depends on accurate evaluation of skin smoothing, blemish removal, texture preservation, and identity consistency.

In this paper, we propose BeautyGRPO, a RL framework that aligns face retouching with human aesthetic preferences. A key challenge in this domain is the lack of reward models capable of assessing fine-grained retouching quality. To address this, we construct FRPref-10K, a large-scale \textbf{f}ace \textbf{r}etouching \textbf{pref}erence dataset covering five key dimensions: skin smoothing, blemish removal, texture quality, clarity, and identity preservation. Each image is evaluated by multiple vision-language models (VLMs) through reasoning-based scoring, followed by human refinement to ensure alignment with aesthetic judgment. Based on this dataset, we train a reward model that captures subtle perceptual differences between retouched image pairs and reflects human preferences.

To resolve the critical conflict between fidelity and exploration, we introduce Dynamic Path Guidance (DPG), the core algorithm of BeautyGRPO. DPG stabilizes the stochastic trajectory by softly guiding it toward an ideal deterministic path while retaining controlled stochasticity. Unlike SFT, which explicitly aligns model outputs with label images through pixel-level supervision, DPG incorporates label images only within the sampling process as stability anchors. These anchors are high-preference exemplars selected from our FRPref-10K dataset and serve to regularize the trajectory around a high-fidelity manifold rather than act as paired ground-truth references. The anchors provide gentle directional stabilization, keeping the sampling trajectory close to a high-fidelity manifold without interfering with the reward-driven optimization. This design allows the policy to move beyond the anchor when higher rewards suggest more appealing results, preserving the exploratory nature of online RL while maintaining trajectory stability. Through this anchored yet reward-driven mechanism, DPG establishes a balanced regime between exploration and fidelity, effectively preventing trajectory drift while enabling the discovery of retouching results that surpass the aesthetic quality of the reference data.

Our contributions are summarized as follows:
\begin{enumerate}
    \item We construct FRPref-10K, a large-scale and fine-grained face retouching preference dataset, and train a specialized reward model that captures multi-dimensional human aesthetic preferences with high accuracy.
    \item We propose BeautyGRPO, a reinforcement learning framework equipped with Dynamic Path Guidance, which effectively balances exploration and fidelity during the retouching process.
    \item Extensive experiments demonstrate that the proposed reward model aligns closely with human aesthetic judgments, and BeautyGRPO achieves superior user preference over existing methods, particularly in skin smoothing, blemish removal, and texture preservation.
\end{enumerate}

\section{Related Work}
\label{sec:related_work}
\subsection{Face Retouching}
Face retouching aims to remove facial imperfections and enhance facial aesthetics while preserving facial identity features.
Early works progress from traditional filtering \cite{arakawa2004nonlinear,batool2014detection,velusamy2020fabsoften} and restoration methods \cite{wang2022restoreformer,wang2023restoreformer++,liu2025mofrr} to CNN-based models \cite{lei2022abpn,xie2023blemish,xu2025face} and more recently to Transformer-based architectures \cite{xue2024vretoucher,wen2024retouchformer,xue2025retouchgpt}. Most of these specialized retouching models still rely on supervised learning.
The limitation lies in their objective to align outputs with label images at the pixel level rather than with human aesthetic preferences. Furthermore, large-scale general image editing models \cite{google2025nanobanana,openai2025gptimage,zhang2025ICEdit,liu2025step1x,wu2025qwenimagetechnicalreport} can perform complex retouching operations through instruction-driven editing, but they often introduce visually unnatural effects such as altered facial identity, oily or plastic-like skin texture, and false visual artifacts, leading to inconsistent and less realistic results.

\subsection{Reinforcement Learning for Diffusion Models}
Reinforcement learning has emerged as an effective approach for aligning diffusion and flow-matching models with human preferences. Policy-gradient methods such as DDPO \cite{DDPO2024} and DPOK \cite{fan2024dpok} optimize preference scores, but the long gradient paths in diffusion models limit updates to only a few late timesteps. DiffusionDPO \cite{wallace2024diffusion} extends DPO \cite{rafailov2023direct} to T2I generation in an offline setting using paired positive and negative samples. This yields stable optimization but remains constrained by the quality and diversity of static data and has limited capacity to go beyond the training distribution. Online RL methods such as FlowGRPO \cite{liu2025flow} adapt GRPO \cite{guo2025deepseekr1} to flow-matching models by converting the deterministic ODE into an SDE to enable active exploration. When applied to face retouching, SDE-based exploration introduces random drift that violates high-fidelity constraints and leads to residual artifacts. While such stochasticity is valuable for diversity in text-to-image generation, it conflicts with the precision and stability required for portrait enhancement.

\subsection{Reward Modeling for Human Preferences}
Reward models are central to reinforcement learning, serving as the optimization signal that aligns model behavior with human preferences. Recent advances in T2I generation and image editing achieve such alignment by training reward models on large-scale human preference datasets \cite{kirstain2023pick,wu2023HPSv2}.
Representative models include ImageReward \cite{xu2023imagereward}, HPSv2 \cite{wu2023HPSv2}, and UnifiedReward \cite{wang2025unified} for T2I generation, and EditScore \cite{luo2025editscore}, EditReward \cite{wu2025editreward}, and UnifiedReward-Edit \cite{wang2025unified} for image editing. While these reward models align generation with human aesthetic judgments, they focus mainly on global image quality and text-image consistency.
They fail to capture subtle perceptual dimensions such as the balance between skin smoothness and texture naturalness, and the cleanliness of blemish removal.
To bridge this gap, we construct a dedicated face retouching preference dataset and train a reward model that provides fine-grained reward guidance aligned with human aesthetic preferences.

\begin{figure*}[htbp] 
\begin{center}
\includegraphics[width=\textwidth]{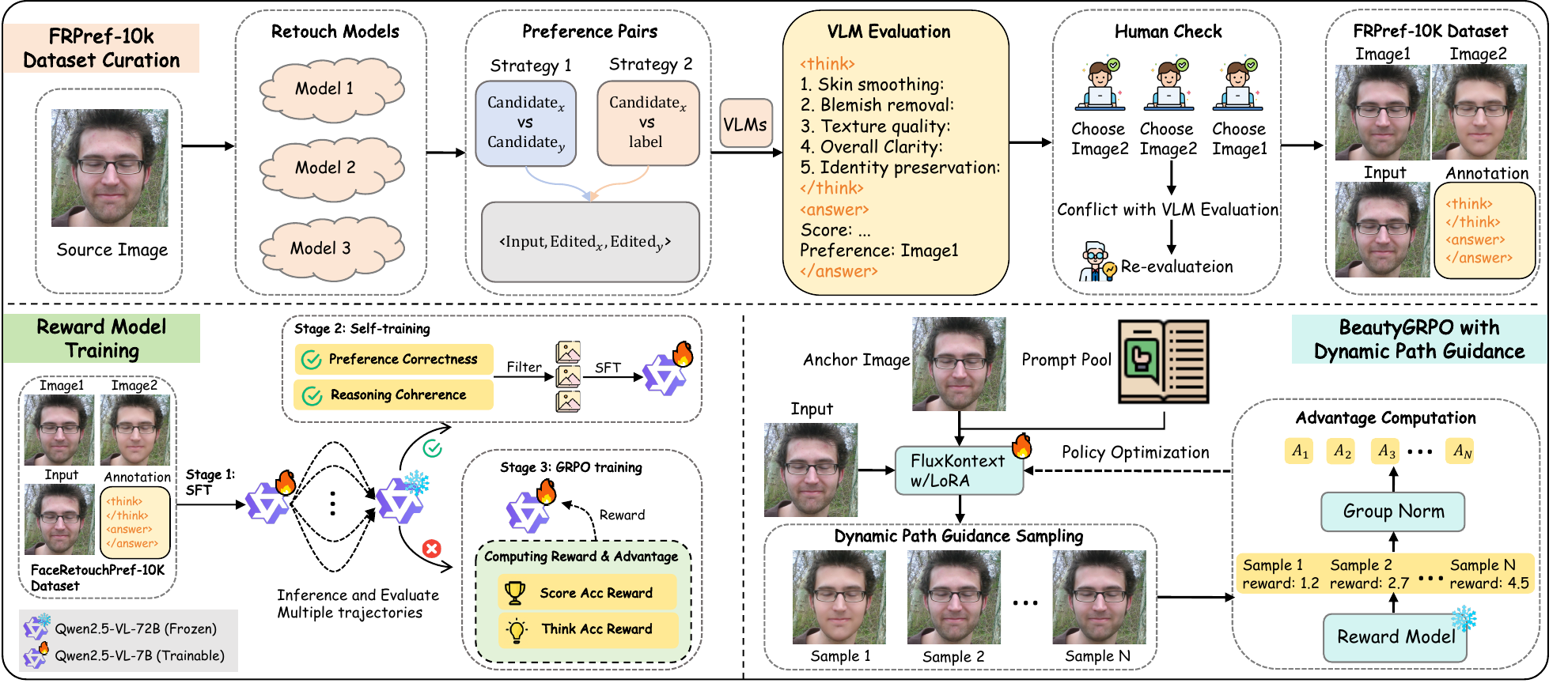}
\end{center}
\caption{Overview of our FRPref-10K construction, reward model training, and BeautyGRPO with Dynamic Path Guidance (DPG). \textbf{Top:} FRPref-10K dataset curation pipeline. Multiple retouched candidates are generated with diverse editing models, preference pairs are formed via output-vs-output/label comparisons, and are annotated by VLMs across five quality dimensions before human verification.
\textbf{Bottom left:} Three-stage reward model training, including SFT, self-training with consistency filtering, and GRPO. 
\textbf{Bottom right:} BeautyGRPO training with DPG on a FluxKontext-LoRA backbone.}
\label{datapipeline}
\end{figure*}

\section{Methodology}
We introduce BeautyGRPO, a reinforcement learning framework that aligns face retouching models with human aesthetic preferences. It consists of three components: (1) a large-scale, multi-dimensional preference dataset (Section~\ref{sec:dataset}); (2) a specialized reward model for fine-grained aesthetic evaluation (Section~\ref{sec:reward_model}); and (3) the Dynamic Path Guidance (DPG) algorithm, which stabilizes online RL by reconciling stochastic exploration with deterministic fidelity (Section~\ref{sec:beautygrpo}).

\subsection{Fine-Grained Preference Data Curation}
\label{sec:dataset}

Aesthetic alignment requires data that reflect the nuanced judgments inherent in face retouching. To overcome the limitations of existing coarse datasets, we construct FRPref-10K, a fine-grained dataset of 10,000 high-resolution preference pairs with detailed multi-dimensional annotations.

\textbf{Data Collection and Candidate Generation.} We curated source portraits from the FFHQR~\cite{Shafaei_2021_ffhqr} and a proprietary high-resolution collection, ensuring diverse demographics and imaging conditions. For each input image $I_{in}$, we generate a diverse pool of retouched candidates $\{I_{out}^i\}$ using multiple state-of-the-art models (NanoBanana~\cite{google2025nanobanana}, Flux.1 Kontext~\cite{fluxkontext2025}, RetouchFormer~\cite{wen2024retouchformer}) with varied random seeds. Preference pairs $(I_{in}, I_A, I_B)$ are then constructed through both inter-model comparison (output vs. output) and output-vs-label evaluation. More details regarding the dataset statistics are provided in the supplementary material~\ref{supp:dataset_composition}.

\textbf{Hybrid VLM-Human Annotation Pipeline.} We employ a three-stage annotation protocol that combines the reasoning ability of large VLMs with human verification:
\begin{enumerate}
    \item \textbf{Multi-Dimensional Assessment:} We define five critical aesthetic dimensions: Skin Smoothing, Blemish Removal, Texture Quality, Clarity, and Identity Preservation. An ensemble of VLMs (GPT-4o~\cite{openai2024gpt4ocard}, Qwen2.5-VL-72B~\cite{bai2025qwen25vl}, Gemini 2.5 Pro~\cite{comanici2025gemini}) evaluates each pair across these dimensions, providing both quantitative scores and structured reasoning. VLM outputs are aggregated to form an initial consensus. Trained human annotators subsequently audit these assessments under identical criteria.
    \item \textbf{Expert Adjudication:} Discrepancies between human judgment and the VLM consensus are escalated to senior experts for final arbitration.
\end{enumerate}

\subsection{Multi-Dimensional Retouching Reward Model}
\label{sec:reward_model}
Leveraging FRPref-10K, we train a specialized reward model $R_\phi(I_{in}, I_{out})$ capable of providing dense, interpretable feedback. Following the UnifiedReward-Thinking paradigm~\cite{wang2025unifiedthinking}, we adopt a three-stage training strategy to enhance reasoning quality, initializing from Qwen2.5-VL-7B~\cite{bai2025qwen25vl}.

\textbf{Stage 1: Structured Reasoning Initialization}
We perform SFT on a 2K subset to adapt the model to generate structured reasoning across five dimensions in a \texttt{<think>} block and a final preference decision in an \texttt{<answer>} block.

\textbf{Stage 2: Self-Training with Consistency Filtering}
The Stage 1 model generates multiple reasoning trajectories for the remaining 8K samples. We employ a consistency filtering mechanism based on two criteria: \textit{Preference Correctness} (alignment with ground truth) and \textit{Reasoning Coherence} (semantic consistency between reasoning and decision, evaluated by a high-capacity VLM). Consistent trajectories are utilized for a subsequent round of SFT, reinforcing reliable reasoning patterns.

\textbf{Stage 3: Robustness Enhancement via GRPO}
For inconsistent samples, we apply GRPO~\cite{guo2025deepseekr1} to improve robustness. The model explores diverse reasoning paths, evaluated by both outcome rewards (preference accuracy) and process rewards (reasoning coherence, assessed by a lightweight verifier such as DeBERTa-V3~\cite{he2023deberta}). This reinforcement phase strengthens the model’s ability to produce stable and human-aligned aesthetic judgments.


\subsection{BeautyGRPO with Dynamic Path Guidance}
\label{sec:beautygrpo}
We use the trained reward model as feedback to align a flow-matching image editing policy model $\pi_\theta$. We extend the FlowGRPO framework~\cite{liu2025flow} with a novel Dynamic Path Guidance (DPG) mechanism that stabilizes SDE-based exploration for high-fidelity retouching tasks.

\subsubsection{Preliminaries: The Fidelity-Exploration Conflict}
Flow-matching models~\cite{lipman2023flow, liu2023rectifiedflow} learn a vector field $v_\theta(x_t, t)$ that transports a noise distribution $p_1(x)$ toward a data distribution $p_0(x)$ through a deterministic ODE trajectory.
To enable exploration for online RL, FlowGRPO~\cite{liu2025flow} converts this ODE into a SDE formulation by injecting noise:
\begin{equation}
dx_t = \left[ v_\theta(x_t, t) + \frac{\sigma_t^2}{2t}\left( x_t + (1-t)v_\theta(x_t, t) \right) \right]dt + \sigma_t d\omega,
\label{flow_sde}
\end{equation}
where $d\omega$ represents standard Brownian motion, and $\sigma_t = \eta \sqrt{\frac{t}{1-t}}$ controls the noise level. 
This stochastic term induces diverse sampling trajectories, allowing the policy to receive richer, more differentiated reward signals. 
Specifically, given $G$ generated samples $\{x_0^i\}_{i=1}^{G}$ with corresponding rewards $R(x_0^i)$, FlowGRPO normalizes these rewards into standardized advantages:
\begin{equation}
\hat{A}^i = \frac{R(x_0^i) - \mathrm{mean}(\{R(x_0^i)\}_{i=1}^{G})}
{\mathrm{std}(\{R(x_0^i)\}_{i=1}^{G})}.
\end{equation}

\begin{figure}[tp]
\centering
\begin{subfigure}[b]{\linewidth}
    \centering
    \includegraphics[width=\linewidth]{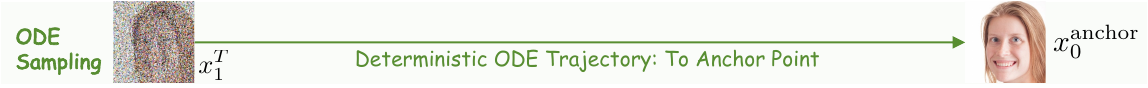}
    \caption{Flow-matching's ODE trajectory to anchor point.}
    \label{fig:method-a}
\end{subfigure}
\vspace{0.6em}
\begin{subfigure}[b]{\linewidth}
    \centering
    \includegraphics[width=\linewidth]{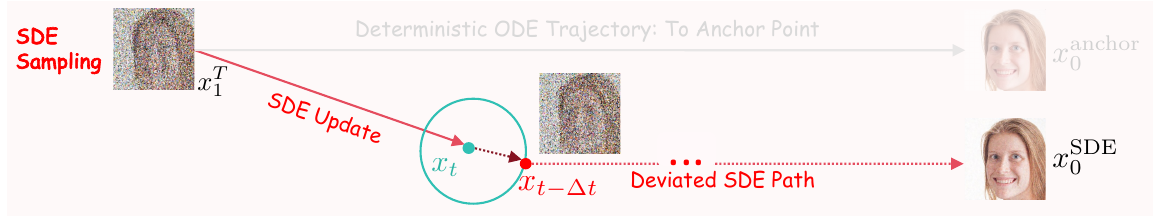}
    \caption{FlowGRPO's SDE sampling trajectory.}
    \label{fig:method-b}
\end{subfigure}
\vspace{0.6em}
\begin{subfigure}[b]{\linewidth}
    \centering
    \includegraphics[width=\linewidth]{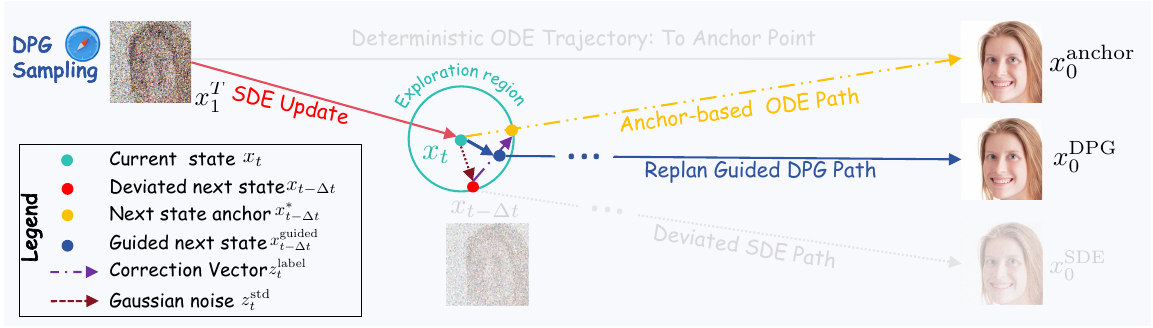}
    \caption{BeautyGRPO with DPG sampling trajectory.}
    \label{fig:method-c}
\end{subfigure}
\caption{Overview of different sampling trajectories. (a) Standard flow-matching ODE trajectory for SFT. (b) Uncontrolled SDE trajectory in FlowGRPO, which gradually drifts from the high-fidelity anchor point and introduces noise artifacts. 
(c) Proposed BeautyGRPO with Dynamic Path Guidance (DPG). At each timestep, DPG dynamically computes an anchor-based ODE path and replans a guided trajectory by linearly blending a correction vector with standard Gaussian noise, correcting stochastic drift while maintaining controlled exploration.}
\label{method}
\end{figure}
The policy is optimized by the GRPO objective:
\begin{align}
\label{eq:grpo_loss}
\mathcal{L}(\theta) = 
\mathbb{E}_{c \sim \mathcal{C}, \{x^i\}_{i=1}^{G} \sim \pi_{\theta_{\text{old}}}} 
\Bigg[ &\frac{1}{G}\sum_{i=1}^{G} \frac{1}{T}\sum_{t=0}^{T-1}
\min\big(r_t^i(\theta)\hat{A}_t^i, \nonumber \\
&\mathrm{clip}(r_t^i(\theta), 1-\varepsilon, 1+\varepsilon)\hat{A}_t^i\big)
\Bigg],
\end{align}
where $r_t^i(\theta) = 
\frac{p_\theta(x_{t-1}^i|x_t^i)}{p_{\theta_{\text{old}}}(x_{t-1}^i|x_t^i)}$ 
denotes the per-step likelihood ratio between old policy model $\pi_{\theta_{\text{old}}}$ and updated model $\pi_\theta$.

While the stochastic term $\sigma_{t}\sqrt{dt}\epsilon$ in Eq.~\ref{flow_sde} enhances exploration in general T2I generation, it conflicts with the high-fidelity demands of face retouching. The sampling trajectory is gradually pushed away from the high-fidelity manifold by the accumulated noise, leading to uncontrolled drift and noticeable visual artifacts.

\subsubsection{Dynamic Path Guidance (DPG) for Anchored Exploration}
To reconcile the conflict between exploration and fidelity, we propose Dynamic Path Guidance (DPG). DPG performs anchored exploration by stabilizing the SDE trajectory near a high-fidelity manifold while still allowing deviations guided by the reward signal.

\textbf{The Stability Anchor.} 
We use high-preference samples from our FRPref-10K dataset as stability anchors, denoted by $x_0^{\text{anchor}}$. The anchor is introduced only in the sampling process and is not treated as a supervision target as in SFT. Its role is to constrain the stochastic trajectory to a high-fidelity manifold, within which the policy can still explore and optimize for higher-reward retouching results.

\textbf{SDE Formulation and Drift.} 
The FlowGRPO SDE update (Eq.~\ref{flow_sde}) can be expressed compactly as:
\begin{equation}
x_{t-\Delta t} = \mu_t + \sigma_{\text{step}}\,z_t, \qquad z_t \sim \mathcal{N}(0,I),
\label{eq:sde-compact}
\end{equation}
where $\mu_t$ is the drift term predicted by the model:
\begin{equation}
\mu_t = x_t + \Big[v_\theta(x_t,t) + \frac{\sigma_t^2}{2t}\big(x_t + (1-t)v_\theta(x_t,t)\big)\Big]\Delta t,
\label{eq:mean}
\end{equation}
and $\sigma_{step} = \sigma_{t}\sqrt{\Delta t}$ is the step-wise standard deviation.
As illustrated in Fig.~\ref{fig:method-b}, repeated application of this update gradually accumulates stochastic drift that drives the trajectory away from the high-fidelity manifold.

\textbf{Dynamic Trajectory Replanning.}
To counteract drift accumulation, DPG dynamically replans a guided  trajectory toward the anchor at each timestep.
As illustrated in Fig.~\ref{fig:method-c}, given the current state $x_t$, we replan a straight line ODE trajectory connecting $x_t$ to the anchor $x_0^{\text{anchor}}$. The corresponding target state at the next timestep $s = t - \Delta t$ on this anchor-guided path is:
\begin{align}
x^{*}_{t-\Delta t}
&= \Big(\frac{\Delta t}{t}\Big)\,x_0^{\text{anchor}}
 + \Big(1-\frac{\Delta t}{t}\Big)\,x_t .
\label{eq:xstar}
\end{align}

\textbf{Guided Correction with Controlled Stochasticity.} We guide the SDE update toward  $x_{t-\Delta t}^{*}$ by introducing a correction vector $z^{\text{anchor}}_t$ that aligns the predicted mean $\mu_t$ with this anchor-guided target:
\begin{equation}
z^{\text{anchor}}_t = \frac{x^{*}_{t-\Delta t}-\mu_t}{\sigma_{\text{step}}}.
\label{eq:z-label}
\end{equation}
This vector indicates the direction that steers the trajectory back toward the anchor-guided path.

However, strictly following $z^{anchor}_t$ would collapse exploration into deterministic imitation, preventing the RL from surpassing the anchor's quality.
To prevent this, we construct a blended noise vector $z^{\text{mix}}_t$ by linearly combining the correction vector with standard Gaussian noise $z_t^{std} \sim \mathcal{N}(0,I)$:
\begin{equation}
z^{\text{mix}}_t = \lambda(t)\,z^{\text{anchor}}_t + \big(1-\lambda(t)\big)\,z^{\text{std}}_t,
\label{eq:z-mix}
\end{equation}
where $\lambda(t)= \frac{t}{\max(1,T-1)}$ is a time-dependent coefficient that decreases along the reverse-time trajectory. 
Thus, early reverse timesteps use stronger anchor guidance to correct structural deviations, while later timesteps rely more on stochastic noise to encourage fine-grained exploration.

\begin{table*}[t]
\centering
\caption{Quantitative evaluation of state-of-the-art methods on the FFHQR and In-the-wild datasets, utilizing diverse no-reference metrics to assess perceptual quality and aesthetic appeal. Our BeautyGRPO significantly outperforms specialized retouching models, general editing baselines, and alternative RL strategies across nearly all benchmarks. These results validate the effectiveness of our preference alignment framework in achieving high-quality, natural retouching outcomes that generalize well across diverse scenarios. Best results are in \textbf{bold}.}
\label{tab:ffhqr-inwild-compact}

\setlength{\tabcolsep}{1pt}
\renewcommand{\arraystretch}{0.99}
\scriptsize

\begin{tabular}{@{}l*{8}{c}|*{7}{c}@{}}
\hline
                                & \multicolumn{8}{c}{FFHQR} & \multicolumn{7}{c}{In-the-wild} \\ \cline{2-16}
                                & NIQE$\downarrow$ & NIMA$\uparrow$ & MUSIQ$\uparrow$ & MANIQA$\uparrow$ & NRQM$\uparrow$ & TOPIQ$\uparrow$ & FID$\downarrow$ & ArcFace$\uparrow$
                                & NIQE$\downarrow$ & NIMA$\uparrow$ & MUSIQ$\uparrow$ & MANIQA$\uparrow$ & NRQM$\uparrow$ & TOPIQ$\uparrow$ & ArcFace$\uparrow$ \\ \hline

\multicolumn{16}{c}{\cellcolor[HTML]{FFE6CD}\textit{Retouch Model}} \\
ABPN\cite{lei2022abpn}                            & 11.152 & 4.861 & 4.599 & 1.027 & 8.219 & 0.611 & 4.713 & 0.971  & 11.979 & 5.005 & 4.651 & 0.997 & 7.019 & 0.603 & 0.968 \\
RestoreFormer\cite{wang2022restoreformer}         & 12.281 & 4.712 & 4.361 & 0.958 & 7.411 & 0.587 & 15.898 & 0.799 & 12.801 & 4.819 & 4.534 & 0.978 & 7.372 & 0.601 & 0.761 \\
RestoreFormer++\cite{wang2023restoreformer++}     & 12.144 & 4.618 & 4.273 & 0.935 & 7.415 & 0.572 & 15.311 & 0.907 & 12.624 & 4.713 & 4.386 & 0.959 & 7.307 & 0.594 & 0.890 \\
VRetouchEr\cite{xue2024vretoucher}                & 12.948 & 4.586 & 4.169 & 0.962 & 6.167 & 0.551 & 2.229  & 0.978 & 13.605 & 4.658 & 4.308 & 0.957 & 5.635 & 0.565 & 0.974 \\
RetouchFormer\cite{wen2024retouchformer}          & 11.153 & 4.723 & 4.465 & 1.036 & 8.175 & 0.605 & 2.275  & \textbf{0.986}  & 12.996 & 4.712 & 4.501 & 0.999 & 6.753 & 0.593 & \textbf{0.976} \\

\multicolumn{16}{c}{\cellcolor[HTML]{CDCDFF}\textit{General Editing Model}} \\
ICEdit \cite{zhang2025ICEdit}                     & 12.502 & 4.645 & 4.192 & 0.978 & 7.424 & 0.555 & 2.818  & 0.967 & 13.016 & 4.175 & 4.371 & 0.967 & 6.610 & 0.567 & 0.949 \\
SeedDream4.0\cite{seedream2025seedream}           & 15.033 & 4.804 & 4.503 & 0.993 & 7.538 & 0.571 & 12.314 & 0.893 & 16.812 & 4.979 & 4.638 & 0.966 & 6.371 & 0.573 & 0.898 \\
NanoBanana\cite{google2025nanobanana}             & 11.301 & 4.919 & 4.681 & 1.009 & 8.065 & 0.621 & 6.804  & 0.889  & 12.043 & 5.131 & 4.783 & 0.989 & 6.935 & 0.632 & 0.881 \\
Flux.K w/LoRA\cite{fluxkontext2025}          & 12.913 & 4.694 & 4.459 & 1.035 & 8.009 & 0.601 & \textbf{1.735} & 0.973 & 12.501 & 4.751 & 4.524 & 0.978 & 6.468 & 0.586 & 0.968 \\

\multicolumn{16}{c}{\cellcolor[HTML]{F4CDDB}\textit{Reinforcement Learning Model}} \\
Flux.K+LoRA w/FlowGRPO\cite{liu2025flow}     & 15.024 & 4.573 & 4.271 & 0.935 & 7.132 & 0.571 & 15.657 & 0.882 & 13.513 & 4.687 & 4.391 & 0.958 & 5.844 & 0.572 & 0.879 \\
Flux.K+LoRA w/Ours                           & \textbf{10.831} & \textbf{5.123} & \textbf{4.906} & \textbf{1.079} & \textbf{8.401} & \textbf{0.676} & 4.054 & 0.952
                                                 & \textbf{11.821} & \textbf{5.357} & \textbf{4.982} & \textbf{1.052} & \textbf{7.526} & \textbf{0.674} & 0.944 \\ \hline
\end{tabular}
\end{table*}

Substituting $z_t^{\text{mix}}$ into the SDE update rule in Eq.~\ref{eq:sde-compact} yields the DPG formulation:
\begin{align}
x^{guided}_{t-\Delta t} 
&= \mu_t + \lambda\,\sigma_{\text{step}}\,z_t^{\text{anchor}} 
        + (1-\lambda)\,\sigma_{\text{step}}\,z_t^{\text{std}} \notag\\
&= (1-\lambda)\,\mu_t + \lambda\,x^{*}_{t-\Delta t}
        + (1-\lambda)\,\sigma_{\text{step}}\,z_t^{\text{std}}.
\label{eq:dpg_update}
\end{align}
which stabilizes online exploration and enables the policy to discover retouching results that achieve higher rewards and better aesthetics than the anchor itself.

\textbf{Log Probability for Policy Optimization.} For the GRPO optimization (Eq.~\ref{eq:grpo_loss}), we require the transition probability $p_{\theta}(x_{t-\Delta t}|x_{t})$ under the guided update rule.
From Eq.~\ref{eq:dpg_update}, this conditional distribution follows a Gaussian $\mathcal{N}(\mu_{\text{new}}, \sigma_\text{new}^2 I)$ where
$\mu_{\text{new}}=(1-\lambda)\,\mu_t+\lambda\,x^{*}_{t-\Delta t}$ and 
$\sigma_{\text{new}}=(1-\lambda)\,\sigma_{\text{step}}$. 
The corresponding log-likelihood is given by:
\begin{align}
\log p_\theta(x_{t-\Delta t} \mid x_t)
&= -\frac{\|x_{t-\Delta t}-\mu_{\text{new}}\|_2^2}{2\,\sigma_{\text{new}}^{2}} -C,
\label{eq:log-prob}
\end{align}
where $C$ denotes constant normalization terms. These per-step probabilities are used to compute the likelihood ratio $r_t(\theta)$ for policy updates.
To reduce computation, we divide the full sampling trajectory into $K=3$ segments, apply DPG update to one randomly selected timestep in each segment, and perform ODE updates for the others.
\section{Experiments}
This section validates BeautyGRPO through quantitative, qualitative, and user study comparisons against state-of-the-art methods, followed by ablation studies on our reward model and the Dynamic Path Guidance (DPG) mechanism. Further experimental details, settings, and results are provided in the supplementary material due to page limitations.

\subsection{Experiments Setup}
\textbf{Training setup.} For all open-source baselines, we perform LoRA \cite{hu2022lora} using editing triplets $\langle\text{prompt}, \text{input}, \text{edited}\rangle$ constructed from our FRPref-10K dataset.
For our main study, we adopt FluxKontext~\cite{fluxkontext2025} as the backbone and apply the same LoRA fine-tuning on these triplets.
Starting from this LoRA-adapted backbone, we conduct RL with FlowGRPO and BeautyGRPO respectively, using the same set of triplets for both.
In each case, the GRPO objective in Eq.~\ref{eq:grpo_loss} is optimized with rewards predicted by our retouching reward model.
The test set contains 1,000 portraits from FFHQR and 1,000 in-the-wild images collected from the Internet. All experiments are run on 8 H20 GPUs.

\textbf{Evaluation metrics.}
Face retouching is inherently subjective, so we adopt no-reference perceptual and aesthetic metrics that approximate human judgment. We use NIQE~\cite{mittal2013niqe} and NRQM~\cite{ma2017nrqm} to assess overall naturalness, NIMA (AVA)~\cite{talebi2018nima}, MUSIQ (AVA)~\cite{ke2021musiq}, MANIQA~\cite{yang2022maniqa}, and TOPIQ (GFIQA)~\cite{chen2024topiq} to measure aesthetic quality, and ArcFace scores to evaluate facial quality and identity preservation. In addition, we report FID on FFHQR to quantify distributional realism. Detailed rationales for metric selection and evaluation specifics are provided in the supplementary material~\ref{sup:metric}.

\begin{figure*}[htbp] 
\begin{center}
\begin{tabular}{c}
\includegraphics[width=\textwidth]{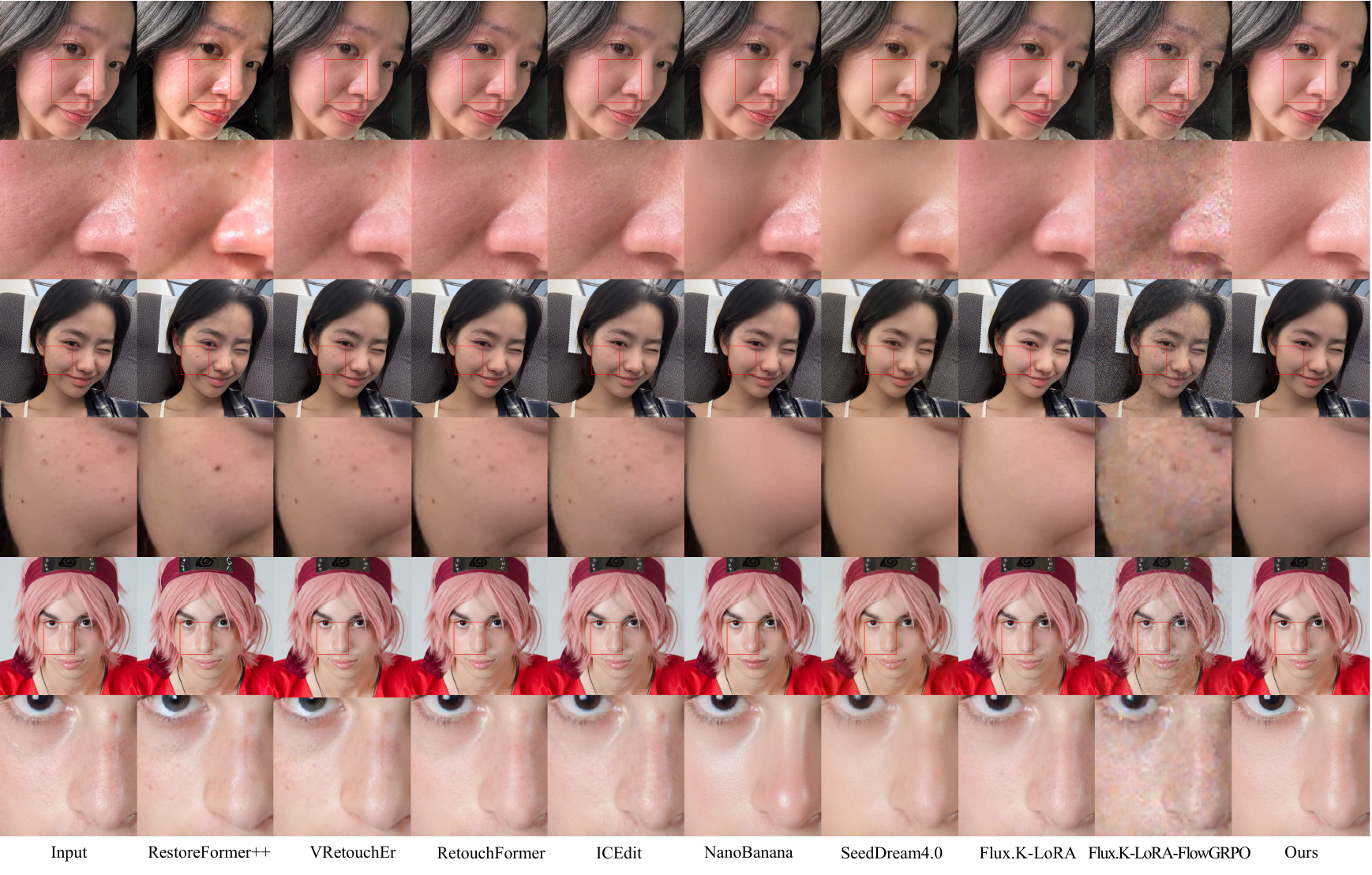}
\end{tabular}
\end{center}
\caption{Visual comparison of face retouching results across different methods on FFHQR and in-the-wild datasets, where Flux.K denotes FluxKontext. 
Our BeautyGRPO removes blemishes cleanly while retaining natural texture, skin gloss, and moles, unlike existing models that exhibit incomplete blemish removal, over-smoothing, or artificial appearance. Please refer to the supplementary materials for more results.}
\label{main_results}
\end{figure*}

\subsection{Main Results}
\subsubsection{Quantitative results}
Table~\ref{tab:ffhqr-inwild-compact} reports the quantitative comparison on FFHQR and the in-the-wild test set.
For face retouching, the quality of a result is ultimately judged by human aesthetic perception when comparing the input and the retouched image, rather than by pixel-level match with a limited set of reference labels.
Full-reference metrics such as PSNR, SSIM, and LPIPS mainly measure pixel fidelity to these references and therefore do not reliably reflect perceived naturalness or visual appeal.
We thus focus on no-reference perceptual and aesthetic metrics that better capture naturalness, realism, and aesthetic quality. As shown in Table~\ref{tab:ffhqr-inwild-compact}, BeautyGRPO outperforms all baselines on both datasets. On FFHQR, it achieves the best scores on all six no-reference metrics, with clear gains in NIMA, MUSIQ, MANIQA, NRQM, and TOPIQ and a noticeably lower NIQE. Furthermore, we evaluate our method on an in-the-wild dataset of 1000 internet portrait images. Similar improvements on this set indicate that these benefits generalize beyond the curated dataset. Although the FID of BeautyGRPO is slightly higher than that of the best supervised baseline, this is expected.
By explicitly optimizing for human preference, BeautyGRPO explores a broader range of plausible retouching styles and moves beyond the dataset-specific label distribution that FID relies on, discovering solutions that better align with human aesthetic judgments than those represented in the training labels. While aesthetic enhancements inherently require texture and structure adjustments that slightly impact identity similarity, our method maintains high ArcFace scores (0.952 on FFHQR, 0.944 in-the-wild) to ensure robust identity preservation.

\subsubsection{Visualization results}
Fig.~\ref{main_results} shows visual comparisons of face retouching results across different methods.
Specialized retouching models like RetouchFormer and VRetouchEr often exhibit inaccurate blemish detection, incomplete blemish removal, and excessive smoothing that erases skin texture.
General editing models like NanoBanana and SeedDream4.0 tend to overedit the face, misidentifying moles as defects and producing an unnaturally glossy, synthetic appearance.
In contrast, our BeautyGRPO removes facial blemishes cleanly while preserving skin texture, pores, natural gloss, and balanced skin tone, and retaining distinctive features such as moles, yielding a natural and visually refined result with excellent identity consistency.

\begin{table}[t]
\centering
\caption{Comparison of user preference win rates for various face retouching methods.}
\label{tab:win_rate}
\setlength{\tabcolsep}{2pt}     
\renewcommand{\arraystretch}{0.99}
\footnotesize
\begin{tabular}{@{}lccccc@{}}   
\hline
         & VretouchEr & RetouchFormer & NanoBanana & Kontext Lora & Ours  \\ \hline
Win rate & 6.50       & 8.50          & 9.75       & 12.00        & \textbf{63.25}  \\ \hline
\end{tabular}
\end{table}

\subsection{User Study}
\begin{figure}[tbp] 
\begin{center}
\includegraphics[width=\linewidth]{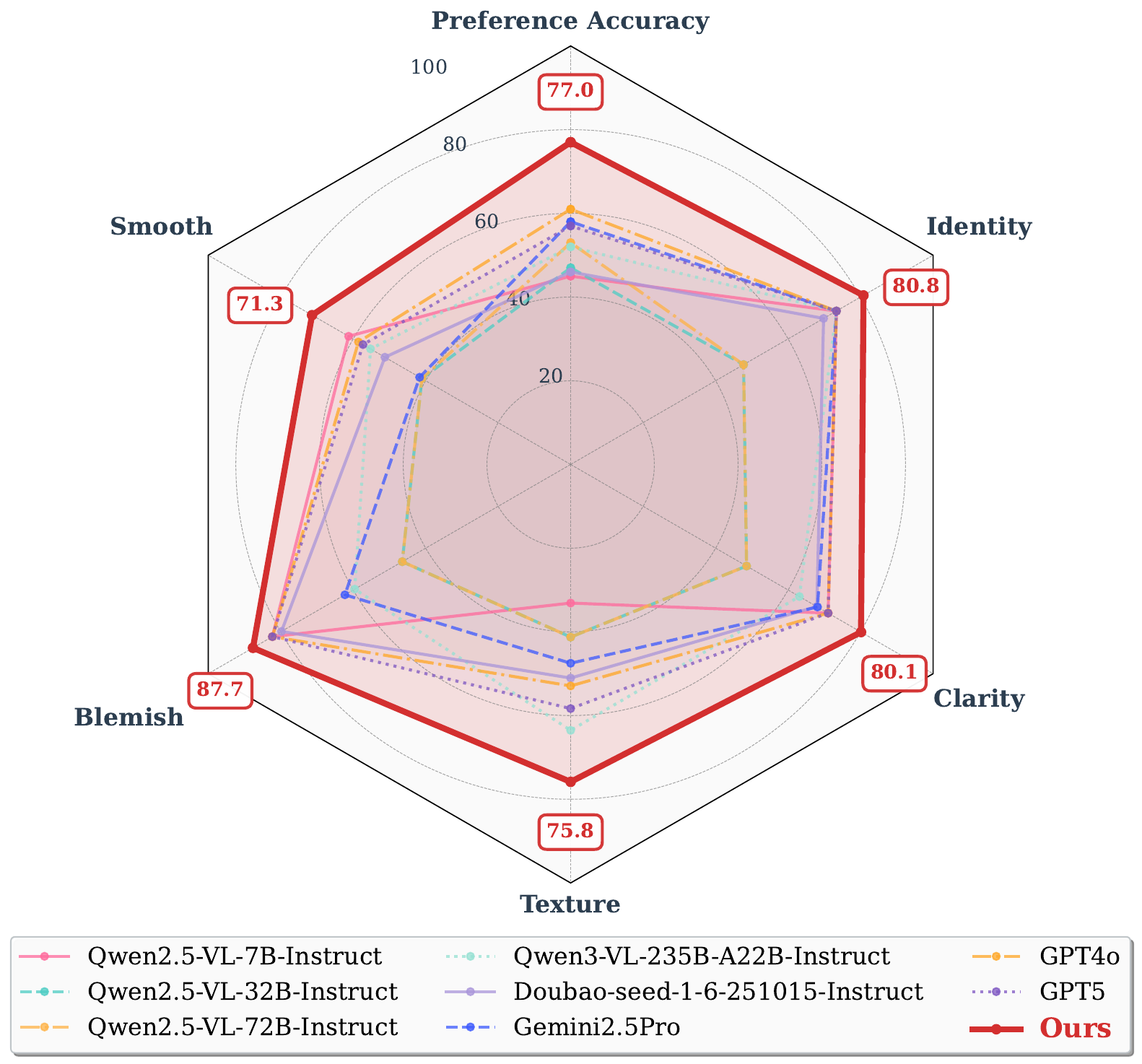}
\end{center}
\caption{Radar chart comparing human judgment accuracy for our reward model against various VLMs. }
\label{survey2_radar}
\end{figure}

We conduct a user study with 100 participants of different ages and editing skill levels on 20 randomly selected test cases. The study contains two questionnaires. In Questionnaire I, each question shows one original portrait and five retouched results from different methods, and participants choose the image they prefer most. In Questionnaire II, each question shows one original image and one retouched result, and participants rate the retouching on five perceptual dimensions, namely skin smoothing, blemish removal, texture quality, clarity, and identity preservation, using a 5-point scale. As shown in Table~\ref{tab:win_rate}, our method achieves clearly higher win rates, indicating stronger alignment with human aesthetic preferences. Fig.~\ref{survey2_radar} compares our reward model with other VLMs in terms of scoring accuracy on the five dimensions and the final preference score, showing that our reward model captures fine-grained retouching preferences more reliably. Further details of the user study setup are provided in the supplementary materials.

\subsection{Ablation study}
\subsubsection{Reward Model}
We integrate BeautyGRPO with various edit reward models on FluxKontext-LoRA, as reported in Table~\ref{tab:abl_rm}. Existing reward models mainly assess coarse image–instruction alignment and global perceptual quality, making them suboptimal for face retouching. In contrast, our reward model provides fine-grained reward modeling on five key dimensions, leading to consistent gains across all metrics.

\subsubsection{BeautyGRPO on Qwen-Image-Edit}
\begin{figure}[tp] 
\begin{center}
\includegraphics[width=\linewidth]{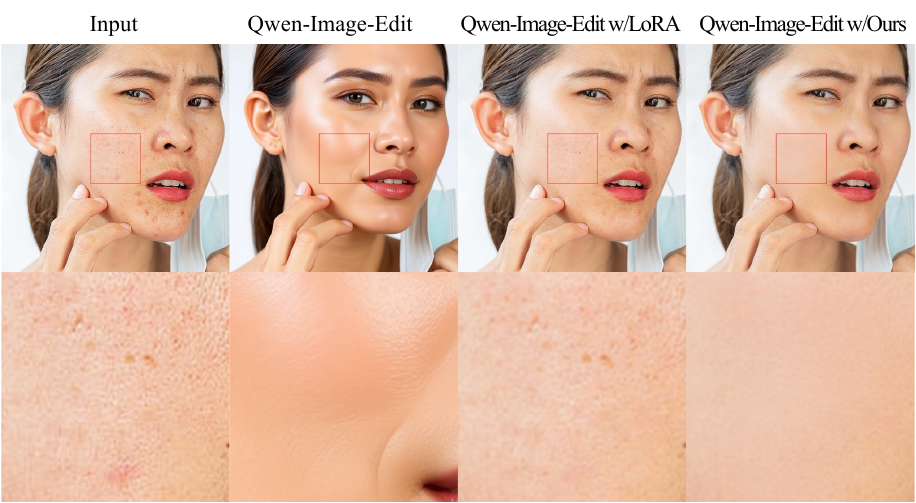}
\end{center}
\caption{Face retouching results of Qwen-Image-Edit with various training paradigms.}
\label{fig:qwen}
\end{figure}

We further evaluate backbone generalization by applying BeautyGRPO to Qwen-Image-Edit~\cite{wu2025qwenimagetechnicalreport}. As shown in Table~\ref{tab:qwen_image_edit}, our BeautyGRPO-trained variant outperforms both the official pretrained and LoRA-tuned models on all metrics. As shown in Fig.~\ref{fig:qwen}, the pretrained model produces unrealistic faces with identity shifts, and the LoRA-tuned model still leaves blemishes and over-smooths skin, whereas BeautyGRPO achieves cleaner blemish removal and more natural skin texture, thereby yielding more realistic and visually pleasing results.

\begin{figure}[tp] 
\begin{center}
\includegraphics[width=1.04\linewidth]{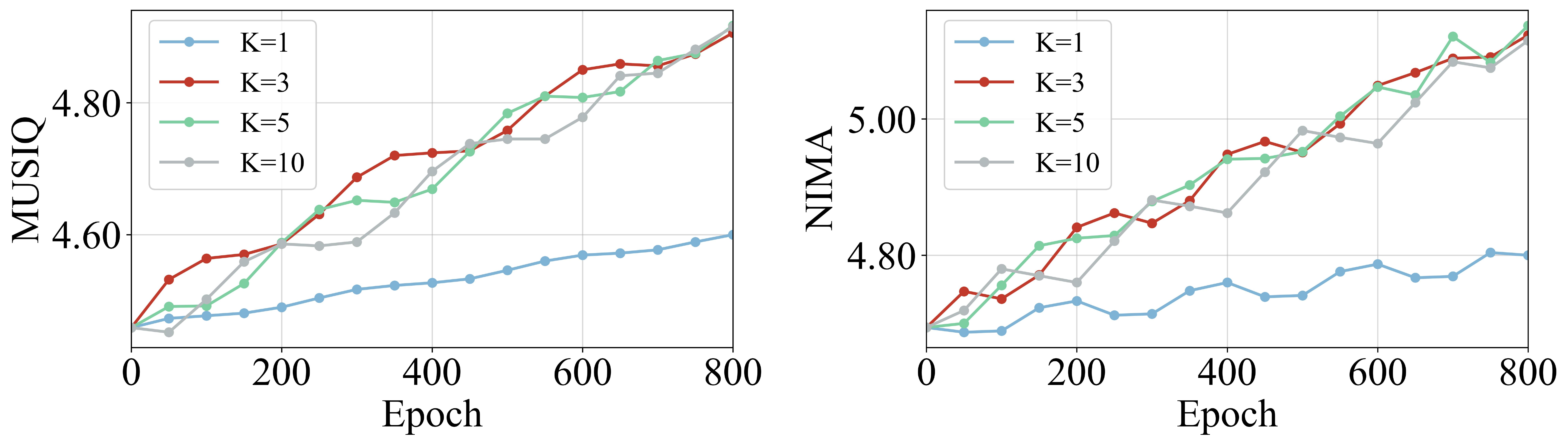}
\end{center}
\caption{Ablation of DPG steps $K$ on FFHQR.}
\label{fig:k}
\end{figure}

\begin{table}[tp]\centering\scriptsize
\setlength{\tabcolsep}{1pt}
\caption{Quantitative evaluation of BeautyGRPO across various reward models on FFHQR.}
\setlength{\tabcolsep}{2pt}     
\renewcommand{\arraystretch}{0.99}
\footnotesize
\label{tab:abl_rm}
\begin{tabular}{lcccc}
\toprule
 & NIMA$\uparrow$ & MUSIQ$\uparrow$ & MANIQA$\uparrow$ & TOPIQ$\uparrow$ \\
\midrule
EditReward\cite{wu2025editreward}             & 4.754 & 4.462 & 1.028 & 0.613 \\
EditScore\cite{luo2025editscore}              & 4.737 & 4.576 & 1.045 & 0.621 \\
UnifiedReward-Edit\cite{wang2025unified}     &4.756 & 4.571 & 1.037 & 0.625 \\
\textbf{Our Reward Model}          & \textbf{5.123} & \textbf{4.906} & \textbf{1.079} & \textbf{0.676}\\
\bottomrule
\end{tabular}
\end{table}

\begin{table}[tp]\centering\scriptsize
\setlength{\tabcolsep}{3pt}
\caption{Quantitative comparison of Qwen-Image-Edit variants, demonstrating the superior performance of integrating BeautyGRPO.}
\label{tab:qwen_image_edit}
\setlength{\tabcolsep}{2pt}     
\renewcommand{\arraystretch}{0.99}
\footnotesize
\begin{tabular}{lcccc}
\hline
                       & NIMA$\uparrow$ & MUSIQ$\uparrow$ & MANIQA$\uparrow$ & TOPIQ$\uparrow$ \\ \hline
Qwen-Image-Edit        & 4.571 & 4.535  & 0.991 & 0.563 \\
Qwen-Image-Edit+LoRA & 4.824 &   4.745 & 1.028 & 0.595 \\
Qwen-Image-Edit+LoRA w/Ours & \textbf{5.351} & \textbf{5.133} & \textbf{1.075}  & \textbf{0.664} \\ \hline
\end{tabular}
\end{table}

\subsubsection{Different Guided SDE Step $K$}
We conduct an ablation study on the number of Dynamic Path Guidance (DPG) steps $K$ used during sampling. As shown in Fig.~\ref{fig:k}, the MUSIQ and NIMA scores at $K=3$ are comparable to those at $K=5$, while requiring fewer DPG updates and thus enabling faster sampling. Therefore, we adopt $K=3$ as the default setting.

\section{Conclusion}
We present BeautyGRPO, an RL-based approach designed to align face retouching with human aesthetic preferences under high-fidelity constraints. We construct the FRPref-10K dataset and train a reward model that performs fine-grained reward modeling of face retouching quality. In addition, we introduce DPG to reconcile stochastic exploration with stable, high-fidelity sampling trajectories. Extensive experiments and user studies demonstrate that BeautyGRPO produces high-quality retouching results that are more consistent with human aesthetic preferences.

\section*{Acknowledgments}
The authors would like to thank the support of Guangdong Provincial Key Laboratory of Information Security Technology (Grant No.2023B1212060026), Guangdong Provincial Key Laboratory of Intelligent Information Processing and Shenzhen Key Laboratory of Media Security (Grant No.2023B1212060076). This work was completed during the first author's internship at vivo BlueImage Lab, vivo Mobile Communication Co.,Ltd., China. 

{
    \small
    \bibliographystyle{ieeenat_fullname}
    \bibliography{main}
}

\clearpage
\appendix
\setcounter{page}{1}
\maketitlesupplementary

\section{Mathematical Derivations for Dynamic Path Guidance}
\subsection{From Flow-Matching ODE to SDE (FlowGRPO)}
\label{sec:flowgrpo_sde}

\paragraph{Flow-Matching ODE.}
In flow-matching generative models, a deterministic ordinary differential equation (ODE) is used to transport a simple noise distribution $p_1(x)$ at time $t=1$ to the data distribution $p_0(x)$ at $t=0$ along a continuous trajectory.
The model learns a time-dependent vector field $v_\theta(x,t)$ such that integrating
\begin{equation}
dx_t = v_\theta(x_t,t)\,dt
\end{equation}
evolves $x_t$ from pure noise at $t=1$ to a realistic sample $x_0$ at $t=0$.
Strictly following this learned ODE trajectory typically yields high-fidelity images.
However, a deterministic path is incompatible with exploration in a reinforcement learning (RL) setting: we need stochasticity to explore alternative outcomes and obtain diverse reward signals.

\paragraph{ODE-to-SDE conversion (FlowGRPO).}
FlowGRPO~\cite{liu2025flow} converts the deterministic ODE into a stochastic differential equation (SDE) while preserving the marginal distributions of the original ODE at every time $t$.
Specifically, it injects Gaussian noise at each sampling step in exploration process:
\begin{equation}
dx_t
=
\Big[
v_\theta(x_t,t)
+
\frac{\sigma_t^2}{2t}\big(x_t + (1-t)v_\theta(x_t,t)\big)
\Big]dt
+
\sigma_t\,d\omega_t,
\label{supp:flowgrpo-sde}
\end{equation}
where $d\omega_t$ denotes the increment of a standard Wiener process, which can be approximated as $d\omega_t \approx \sqrt{dt}\,\epsilon_t$, with $\{\epsilon_t\}$ i.i.d.\ $\mathcal{N}(0,I)$ across time steps, and $\sigma_t = \eta\sqrt{\tfrac{t}{1-t}}$ controls the noise magnitude at time $t$.

Here the SDE in Eq.~\eqref{supp:flowgrpo-sde} is formulated in reverse time (from $t=1$ to $t=0$).
For a small timestep $\Delta t$, the Euler--Maruyama discretization of Eq.~\eqref{supp:flowgrpo-sde} yields the conditional mean
\begin{equation}
\mu_t
=
x_t
+
\Big[
v_\theta(x_t,t)
+
\frac{\sigma_t^2}{2t}\big(x_t + (1-t)v_\theta(x_t,t)\big)
\Big]\Delta t,
\label{eq:mean}
\end{equation}
with a step-wise standard deviation
\begin{equation}
\sigma_{\text{step}} = \sigma_t\sqrt{\Delta t}.
\end{equation}
The one-step update can therefore be written as
\begin{equation}
x_{t-\Delta t}
=
\mu_t + \sigma_{\text{step}} z_t,
\qquad
z_t \sim \mathcal{N}(0,I),
\label{supp:sde-step}
\end{equation}
where $z_t$ is a standard normal random vector.
Eqs.~\eqref{eq:mean}–\eqref{supp:sde-step} show that each update is centered at the model prediction $\mu_t$ and then perturbed by isotropic Gaussian noise, so iterating this stochastic update produces a distribution over possible final states $x_0$ rather than a single deterministic outcome, which in turn enables the policy to explore multiple image variants under the same condition and obtain diverse reward feedback in the RL setting.

\subsection{Fidelity--Exploration Conflict}
While the SDE in FlowGRPO provides the stochasticity needed for online RL, it conflicts with the high-fidelity requirements of face retouching.
From the discrete update in Eq.~\ref{supp:sde-step}, each noise sample $z_t$ perturbs the state away from the model-predicted mean $\mu_t$, and the cumulative term
\[
\sum_t \sigma_{\text{step}}(t)\, z_t
\]
over $T$ sampling steps (we discretize the exploration process into $T$ steps) gradually drives the trajectory $\{x_t\}$ off the high-quality image manifold.
Here $\sigma_{\text{step}}(t) = \sigma_t \sqrt{\Delta t}$ denotes the step-wise standard deviation at time $t$.
In generic text-to-image generation this drift can be tolerable or even helpful for diversity, but in face retouching it mainly appears as noise artifacts.
Reducing $\sigma_t$ mitigates drift but also suppresses exploration, limiting the ability of RL to find higher-reward solutions.
We therefore introduce a mechanism that keeps the trajectory close to a high-fidelity manifold while still enabling controlled stochastic exploration.

\subsection{Dynamic Path Guidance (DPG)}
\label{supp:subsec:dpg}

To resolve the fidelity--exploration conflict, we introduce Dynamic Path Guidance (DPG), which performs anchored exploration by reshaping the noise in the FlowGRPO SDE so that the sampling trajectory remains close to a high-fidelity reference path while still preserving stochasticity.

\textbf{Anchor-guided target.}
We assume access to a high-quality anchor sample $x_0^{\text{anchor}}$ that lies on the desired high-fidelity manifold (e.g., a top-rated retouched face from our preference dataset).
At sampling time $t$ with current state $x_t$, we construct a straight-line path connecting $x_t$ to $x_0^{\text{anchor}}$ over the remaining sampling-time horizon.
The corresponding target at the next time $t-\Delta t$ on this anchor-guided path is defined as
\begin{equation}
x^{*}_{t-\Delta t}
=
\Big(1-\frac{\Delta t}{t}\Big)x_t
+
\frac{\Delta t}{t}\,x_0^{\text{anchor}},
\label{supp:xstar}
\end{equation}
which is a linear interpolation that moves a fraction $\Delta t/t$ of the distance from $x_t$ toward $x_0^{\text{anchor}}$.
Geometrically, $x^{*}_{t-\Delta t}$ lies on the segment joining $x_t$ and $x_0^{\text{anchor}}$ and specifies an anchor-consistent reference direction for the next update.
In practice, $t$ is discretized into $T$ sampling timesteps in the exploration process and we only apply the anchor-guided update for $t>0$, so the division by $t$ is well-defined and the final step at $t=0$ is treated as the terminal state.

\textbf{Anchor correction noise.}
Recall that the FlowGRPO discretization in Eq.~\eqref{supp:sde-step} can be written as
\begin{equation}
x_{t-\Delta t}
=
\mu_t + \sigma_{\text{step}} z_t,
\qquad
z_t \sim \mathcal{N}(0,I),
\end{equation}
where $\mu_t$ is given by Eq.~\eqref{eq:mean} and $\sigma_{\text{step}} = \sigma_t\sqrt{\Delta t}$.
Conditioned on $x_t$, this induces the Gaussian transition
\begin{equation}
x_{t-\Delta t}\mid x_t \sim \mathcal{N}\big(\mu_t,\;\sigma_{\text{step}}^2 I\big).
\end{equation}
The noise that maps the next state exactly to the anchor-guided target $x^{*}_{t-\Delta t}$ in Eq.~\eqref{supp:xstar} is obtained by enforcing
\begin{equation}
 x^{*}_{t-\Delta t} = \mu_t + \sigma_{\text{step}} z_t,
\end{equation}
which yields the anchor correction vector
\begin{equation}
z_t^{\text{anchor}}
=
\frac{x^{*}_{t-\Delta t} - \mu_t}{\sigma_{\text{step}}}.
\label{supp:zanchor}
\end{equation}
The vector $z_t^{\text{anchor}}$ in Eq.~\eqref{supp:zanchor} precisely maps the next state onto the anchor-guided target $x^{*}_{t-\Delta t}$.
However, if this correction is applied deterministically at every step, i.e., $z_t = z_t^{\text{anchor}}$ for all $t$, then $x_{t-\Delta t} = x^{*}_{t-\Delta t}$ holds throughout the sampling process, causing the trajectory to collapse onto the deterministic anchor path in Eq.~\eqref{supp:xstar} and leaving no space for stochastic exploration.

\textbf{Mixed noise and time-dependent guidance.}
To retain stochasticity, DPG replaces the original standard Gaussian $z_t$ with a time-dependent mixture of the anchor correction and an independent standard normal sample.
Let $z_t^{\text{std}} \sim \mathcal{N}(0,I)$ be independent of $x_t$ and define
\begin{equation}
z_t^{\text{mix}}
=
\lambda_t\,z_t^{\text{anchor}}
+
\big(1-\lambda_t\big)\,z_t^{\text{std}},
\quad
0 \le \lambda_t \le 1,
\label{supp:zmix}
\end{equation}
where $\lambda_t$ is a time-dependent guidance weight.
In practice, we discretize the reverse-time sampling process into $T$ steps indexed by integers $t\in\{0,\dots,T-1\}$, where $T$ denotes the total number of steps in the exploration process.
We adopt the same linear schedule as in the main text,
\begin{equation}
\lambda_t = \frac{t}{\max(1,T-1)}.
\end{equation}
Sampling proceeds in reverse time from $t=T-1$ (early, high-noise steps) down to $t=0$ (late refinement), so the effective guidance weight $\lambda_t$ is large at early sampling timesteps and gradually decreases to $0$ toward the final refinement steps.
For notational simplicity, we continue to write $\lambda_t$ in the derivations below to denote this time-dependent weight.
We then replace $z_t$ in the discretization above by $z_t^{\text{mix}}$ and obtain the DPG transition
\begin{equation}
x_{t-\Delta t}
=
\mu_t + \sigma_{\text{step}} z_t^{\text{mix}}.
\label{supp:dpg-update}
\end{equation}

Conditioned on $x_t$, the anchor correction $z_t^{\text{anchor}}$ is a deterministic function of $(x_t, x_0^{\text{anchor}})$ and the only stochastic component is $z_t^{\text{std}}$.
Taking conditional expectation with respect to $z_t^{\text{std}}$ and using $\mathbb{E}[z_t^{\text{std}}]=0$, we obtain
\begin{align}
\mathbb{E}[x_{t-\Delta t}\mid x_t]
&=
\mu_t + \sigma_{\text{step}}\mathbb{E}[z_t^{\text{mix}}\mid x_t] \\
&=
\mu_t + \sigma_{\text{step}}\lambda_t z_t^{\text{anchor}} \\
&=
\mu_t + \lambda_t\big(x^{*}_{t-\Delta t}-\mu_t\big) \\
&=
\big(1-\lambda_t\big)\mu_t + \lambda_t x^{*}_{t-\Delta t}.
\label{supp:dpg-mean}
\end{align}
Thus, in expectation, DPG interpolates the transition mean between the original FlowGRPO mean $\mu_t$ and the anchor-guided target $x^{*}_{t-\Delta t}$, with interpolation weight $\lambda_t$.

Similarly, still conditioning on $x_t$ and treating $z_t^{\text{anchor}}$ as deterministic, the covariance of $z_t^{\text{mix}}$ is
\begin{equation}
\mathrm{Cov}\big[z_t^{\text{mix}}\mid x_t\big]
=
(1-\lambda_t)^2 I,
\end{equation}
since $z_t^{\text{anchor}}$ has zero conditional variance and $\mathrm{Cov}[z_t^{\text{std}}]=I$.
This implies
\begin{equation}
x_{t-\Delta t}\mid x_t
\sim
\mathcal{N}\Big(
\big(1-\lambda_t\big)\mu_t + \lambda_t x^{*}_{t-\Delta t},\;
\big((1-\lambda_t)\sigma_{\text{step}}\big)^2 I
\Big).
\label{supp:dpg-gaussian}
\end{equation}
When $\lambda_t=0$, Eq.~\eqref{supp:dpg-gaussian} reduces exactly to the original FlowGRPO kernel with mean $\mu_t$ and variance $\sigma_{\text{step}}^2 I$.
When $\lambda_t=1$, the variance collapses to zero and the update becomes a deterministic step to $x^{*}_{t-\Delta t}$.
For intermediate values $0<\lambda_t<1$, the trajectory is biased toward the anchor path while retaining a reduced but nonzero variance.

\textbf{Effect on fidelity and exploration.}
Eqs.~\eqref{supp:dpg-mean} and~\eqref{supp:dpg-gaussian} show that DPG simultaneously (i) pulls the mean transition toward an anchor-consistent direction and (ii) shrinks the step variance by a factor of $(1-\lambda_t)^2$.
By choosing $\lambda_t$ large when $t$ is large and the state is still highly noisy, the sampling trajectory in the exploration process is tightly constrained around the high-fidelity anchor manifold during the early, structure-forming stages.
As $t\to 0$ and $\lambda_t\to 0$, the variance gradually recovers to its original level, allowing finer stochastic exploration on top of an already well-structured face.
In this way, DPG reconciles the fidelity requirements of portrait retouching with the exploration needs of online reinforcement learning, purely through a sampling-time modification of the FlowGRPO transition kernel without changing the underlying flow-matching policy $v_\theta$.

\section{Anchored Exploration with Standard GRPO Optimization}

It is important to emphasize that the anchor image is employed exclusively during the sampling-based exploration phase that generates trajectories for GRPO, and it is never used as a supervision target in the optimization objective.
During the exploration phase, for each input portrait in FRPref-10K, we select a high-quality retouched image as the anchor $x_0^{\text{anchor}}$ solely to construct the DPG-guided sampling path. Specifically, at each sampling timestep $t$, DPG computes the anchor-guided target $x^{*}_{t-\Delta t}$ and the corresponding correction noise $z_t^{\text{anchor}}$ (Eqs.~\eqref{supp:xstar}--\eqref{supp:zanchor}), forms the mixed noise $z_t^{\text{mix}}$ in Eq.~\eqref{supp:zmix}, and updates $x_{t-\Delta t}$ via Eq.~\eqref{supp:dpg-update}.
This anchored exploration only changes how trajectories $\{x_t\}$ are sampled during RL rollouts and does not add any reconstruction or regression term toward the anchor in the loss.

In the optimization phase, we apply the standard GRPO objective to the policy model $\pi_\theta$.
The reward model scores the final retouched outputs $x_0$, and these scores are converted into advantages to optimize the policy via the clipped GRPO loss. This optimization process is entirely independent of the anchor image.
In particular, the anchor image $x_0^{\text{anchor}}$ is never used as a pixel-level or feature-level target, and there is no supervised alignment between the policy output and the anchor.
At inference time, DPG is discarded entirely, and we utilize standard ODE sampling. This ensures that the inference process requires no reference images and incurs no additional computational cost.

\section{Relation to Reference-Guided Methods, CFG, and Path Regularization}

Although DPG uses an anchor image during the exploration phase, its role and implementation are fundamentally different from typical reference-guided editing, classifier-free guidance (CFG), or path regularization.

\textbf{Distinction from Reference-Guided Editing.}
Conventional reference-based methods explicitly inject the reference image into the model architecture as an additional conditioning signal (e.g., via concatenation, cross-attention, or style codes), often training the network to strictly imitate the reference appearance through supervised losses.
In contrast, our anchor image is never passed through the retouching network as an input condition and never appears in the optimization objective.
Instead, it serves purely as a geometric anchor to define an anchor-based ODE path in the latent space and to compute the correction noise $z_t^{\text{anchor}}$ within the sampler.
The policy model $\pi_\theta$ remains conditioned solely on the original input portrait and textual prompt, learning aesthetic alignment from reward signals rather than through direct supervision against the anchor.

\textbf{Distinction from Classifier-Free Guidance (CFG).}
CFG modifies the sampling dynamics by adding a guidance term proportional to the difference between conditional and unconditional model outputs. This effectively steers the generation toward regions with higher likelihood under the condition.
In contrast, DPG operates without an external classifier or dual-forward passes. Instead of altering the vector field, DPG reshapes the stochastic transition kernel by linearly mixing a deterministic anchor correction with standard Gaussian noise via a time-dependent weight $\lambda_t$.
This simultaneously biases the transition mean toward the anchor-guided direction (Eq.~\eqref{supp:dpg-mean}) and shrinks the variance (Eq.~\eqref{supp:dpg-gaussian}), keeping the trajectory on a high-fidelity manifold during early sampling while still allowing exploration.
All of this is achieved without requiring auxiliary guidance networks or modifying the drift prediction.

\textbf{Distinction from Path Regularization.}
Path regularization methods typically enforce that the learned vector field or sampling trajectory stay close to a predefined reference path by adding explicit regularization terms to the training loss, penalizing deviations over the entire trajectory.
Our DPG does not add any path penalty or consistency term to the loss function.
The anchor-based ODE path is used only at sampling time to construct $x^{*}_{t-\Delta t}$ and $z_t^{\text{anchor}}$, and the influence of the anchor is further modulated by the schedule $\lambda_t$ that decays along the sampling timesteps.
Thus, DPG acts as a sampling-time constraint on exploration rather than a training-time path regularizer. It regularizes the rollout distribution from which GRPO collects experience, while the GRPO objective itself remains unchanged.

\textbf{Summary.} DPG distinguishes itself from these methods in that: (i) the anchor image is neither a supervised target nor an explicit conditioning input; (ii) no auxiliary guidance network or classifier is employed; and (iii) no regularization terms are added to the training loss.
Instead, DPG is a lightweight exploration-time mechanism that reshapes the SDE sampling kernel using anchor-induced geometry, enabling high-fidelity yet exploratory RL updates while keeping inference-time sampling identical to standard ODE-based flow matching.

\begin{table}[t]
\footnotesize
\setlength{\tabcolsep}{2pt}
\caption{Distortion vs. perceptual quality metrics on FFHQR.}
\label{appendix:rebuttal_metrics}
\centering
\renewcommand{\arraystretch}{0.85}
\begin{tabular}{@{}cl cccccc@{}}
\toprule
& Method & PSNR$\uparrow$ & SSIM$\uparrow$ & LPIPS$\downarrow$ & MUSIQ$\uparrow$ & MANIQA$\uparrow$ & NRQM$\uparrow$ \\ 
\midrule
\multirow{3}{*}[2pt]{\rotatebox[origin=c]{90}{\tiny\textsc{Supervised}}} 
& RetouchFormer & 42.088 & \textbf{0.997} & 0.024 & 4.465 & 1.036 & 8.175 \\ 
& Xu (2025) \cite{xu2025face} & 45.480 & 0.993 & 0.008 & -- & -- & -- \\
& RetouchGPT & \textbf{46.210} & -- & \textbf{0.002} & -- & -- & -- \\
\midrule
\multirow{5}{*}{\rotatebox[origin=c]{90}{\tiny\textsc{Generative}}} 
& ICEdit & 32.463 & 0.882 & 0.152 & 4.192 & 0.978 & 7.424 \\
& SeedDream & 25.598 & 0.739 & 0.170 & 4.503 & 0.993 & 7.538 \\
& NanoBanana & 28.841 & 0.824 & 0.106 & 4.681 & 1.009 & 8.065 \\
& Flux.K w/LoRA & 36.001 & 0.956 & 0.063 & 4.459 & 1.035 & 8.009 \\
& \cellcolor[HTML]{EFEFEF}\textbf{Ours} & \cellcolor[HTML]{EFEFEF}33.697 & \cellcolor[HTML]{EFEFEF}0.927 & \cellcolor[HTML]{EFEFEF}0.102 & \cellcolor[HTML]{EFEFEF}\textbf{4.906} & \cellcolor[HTML]{EFEFEF}\textbf{1.079} & \cellcolor[HTML]{EFEFEF}\textbf{8.401} \\ 
\bottomrule
\end{tabular}
\end{table}

\section{Evaluation Metrics}
\label{sec:metrics}

\subsection{Why Not Use Full-Reference Metrics}
\label{sup:metric}
\begin{figure}[t]
\centering
\includegraphics[trim=0 0 0 10pt, clip, width=\columnwidth]{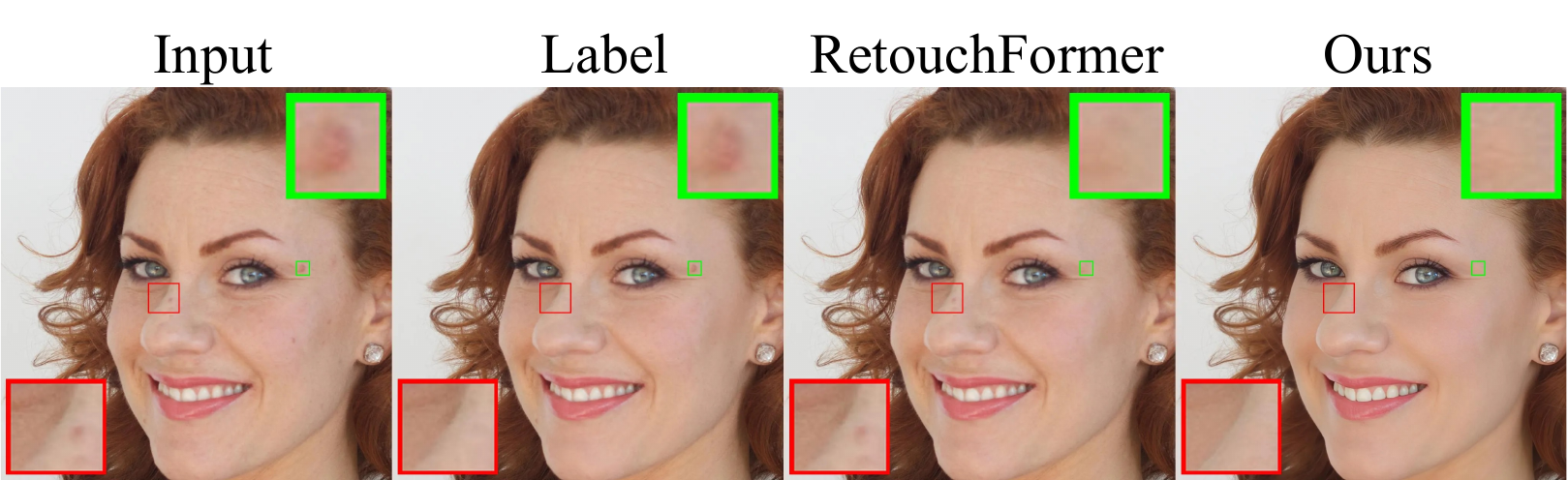}
\caption{Visual examples of imperfect ground-truth references in the FFHQR dataset, which are actively corrected by our method.}
\label{appendix:metric}
\end{figure}

Previous face retouching works~\cite{lei2022abpn,wang2022restoreformer,xue2024vretoucher,wen2024retouchformer} typically adopt classical full-reference (FR) metrics, such as PSNR, SSIM, and LPIPS, to assess pixel-level fidelity between the retouched output and a ground-truth reference. 
However, we maintain that FR metrics are fundamentally ill-suited for evaluating aesthetic enhancement tasks like face retouching, supported by both theoretical and empirical evidence:

\begin{itemize}[leftmargin=*]
    \item \textbf{Theoretical Conflict (The Perception-Distortion Tradeoff).}
    As theoretically justified by Blau and Michaeli~\cite{blau2018perception}, there is an inherent mathematical conflict between maximizing perceptual quality and minimizing pixel-level distortion (e.g., PSNR). Specifically, models strictly minimizing distortion tend to average out high-frequency details, often resulting in overly smoothed and perceptually flat skin textures. In contrast, synthesizing realistic facial details, such as natural pores and lighting variations, inevitably introduces pixel-wise deviations from the reference. This tradeoff is clearly evident in Table~\ref{appendix:rebuttal_metrics}: compared to our supervised baseline (Flux.K w/LoRA), our RL-optimized BeautyGRPO trades pixel fidelity (PSNR dropping from $36.00$ to $33.70$ dB) for significant aesthetic gains (MUSIQ improving from $4.46$ to $4.91$). Optimizing strictly for FR metrics fundamentally restricts a model's ability to achieve peak perceptual quality.

    \item \textbf{Imperfect Ground Truth and Defect Replication.}
    FR metrics implicitly assume that target reference images are flawless. In reality, existing datasets like FFHQR often contain ground-truth references with residual blemishes or suboptimal aesthetic qualities (see Fig.~\ref{appendix:metric}). Models optimizing for high PSNR are therefore forced to replicate these defects to strictly match the reference. In contrast, BeautyGRPO actively corrects these flaws, moving beyond the dataset's inherent limitations to improve naturalness and visual appeal. Consequently, FR metrics unfairly penalize our model for deviating from these flawed references.
\end{itemize}

Consequently, we focus on no-reference (NR) metrics, which directly evaluate naturalness, realism, and aesthetic quality, providing a much more accurate and reliable reflection of true human preference in face retouching.

\subsection{No-Reference Metrics}
For these reasons, we instead focus on no-reference (NR) perceptual and aesthetic metrics that better approximate human judgments in our evaluation.
\begin{itemize}
    \item \textbf{NIQE~\cite{mittal2013niqe} ($\downarrow$).}
    NIQE models the statistical regularities of high-quality natural images and measures how strongly a test image deviates from them.
    For face retouching, it mainly reflects global perceptual naturalness, including whether skin texture appears artificial, overly smooth, or oily, and whether it contains visible AI artifacts.
    A \textbf{lower} score indicates better perceptual naturalness.

    \item \textbf{NIMA~\cite{talebi2018nima} ($\uparrow$).}
    NIMA predicts the distribution of human opinion scores and uses the expected rating as the aesthetic quality measure.
    The AVA-trained model emphasizes overall visual appeal, including composition, color harmony, and aesthetic preference.
    A \textbf{higher} score indicates better aesthetic quality.

    \item \textbf{MUSIQ~\cite{ke2021musiq} ($\uparrow$).}
    MUSIQ provides a holistic assessment of perceptual quality by jointly considering texture fidelity, clarity, and overall aesthetic impression.
    The AVA-trained model is sensitive to global balance as well as fine-grained visual quality, making it suitable for evaluating portrait enhancements.
    A \textbf{higher} score indicates better perceptual quality.

    \item \textbf{NRQM~\cite{ma2017nrqm} ($\uparrow$).}
    NRQM predicts a single perceptual quality score learned from subjective studies.
    In face retouching, it reflects perceived sharpness, texture realism, and local structure fidelity, especially in facial regions.
    A \textbf{higher} score indicates better perceptual quality.

    \item \textbf{MANIQA~\cite{yang2022maniqa} ($\uparrow$).}
    MANIQA is designed to detect complex distortions and generative artifacts.
    It is particularly sensitive to texture integrity, fine facial detail, and subtle artifacts introduced during retouching.
    A \textbf{higher} score indicates better perceptual and textural quality.

    \item \textbf{TOPIQ~\cite{chen2024topiq} ($\uparrow$).}
    TOPIQ follows a top-down procedure that emphasizes distortions in semantically important regions.
    Using the GFIQA-trained face model, it focuses on face-region clarity, texture naturalness, and identity preservation.
    A \textbf{higher} score indicates better face-specific perceptual quality.
\end{itemize}

\section{Composition and Statistics of FRPref-10K}
\label{supp:dataset_composition}
\begin{figure}[tp]
\centering
\includegraphics[width=\columnwidth]{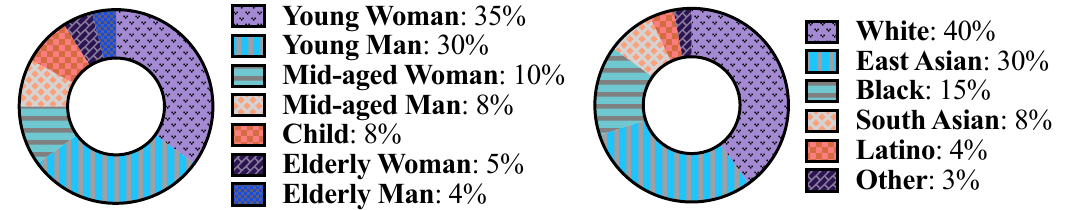}
\caption{Age and racial distribution of the FRPref-10K dataset.}
\label{appendix:dataset}
\end{figure}

To ensure a diverse and representative training corpus for aesthetic alignment, we carefully curated the FRPref-10K dataset. The dataset comprises a mixture of sources: 70\% of the images are sampled from the official training split of the FFHQR dataset, while the remaining 30\% are sourced from a proprietary high-resolution portrait collection. This hybrid approach specifically helps mitigate racial and demographic biases~\cite{maluleke2022studying} present in FFHQR datasets. 
To provide a comprehensive understanding of the dataset's diversity, we detail its demographic distribution in Fig.~\ref{appendix:dataset}.

\section{User Study Details}
\label{sec:supp_user_study}

\subsection{Study Protocol}

We design two complementary questionnaires to evaluate users' perceptual preferences and to assess how well different reward models align with human judgments.

\begin{figure}[tbp] 
\begin{center}
\includegraphics[width=\linewidth]{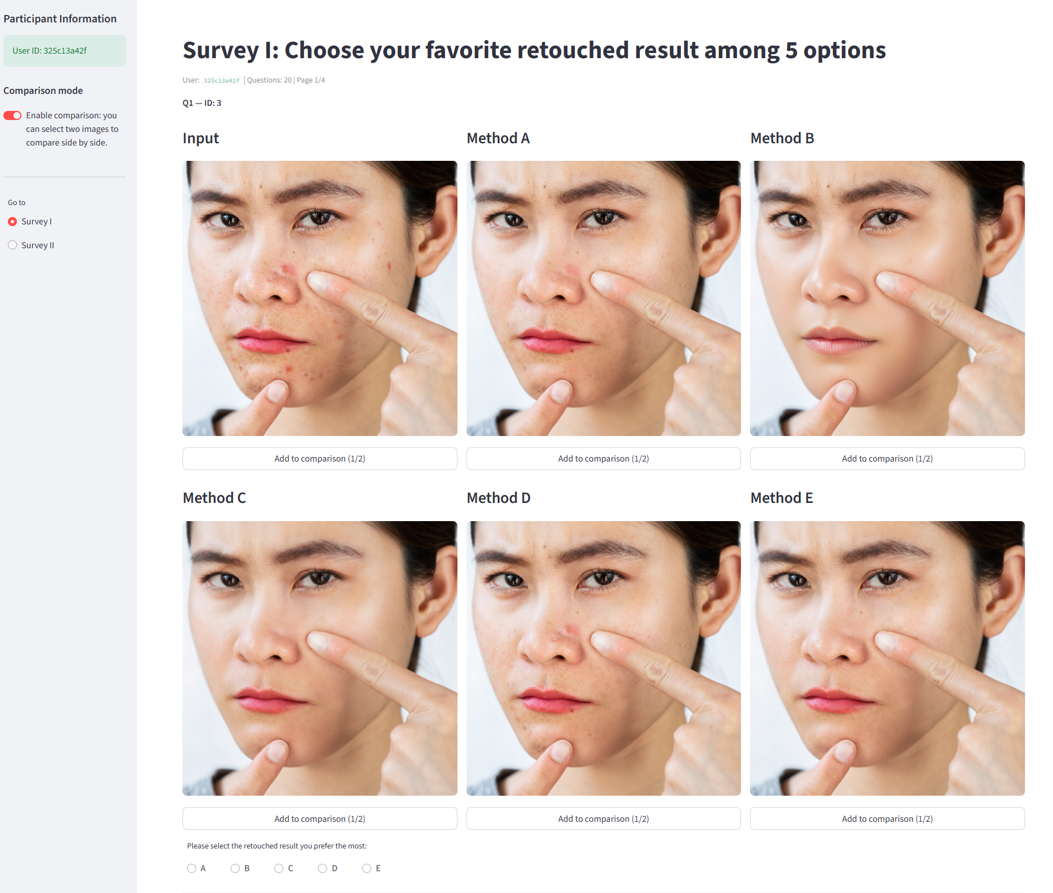}
\end{center}
\caption{User interface of Questionnaire~I.}
\label{supp:survey_1_ui}
\end{figure}

\textbf{Questionnaire I: Overall Preference.}
The first questionnaire consists of a fixed set of 25 questions.
Each question presents one original input portrait and five retouched results produced by different methods, including our approach and representative baselines.
For each question, participants are asked to select one retouched image that they personally prefer the most.
We do not impose any additional instructions or constraints, and users are encouraged to follow their first impression and choose according to their own subjective aesthetic preferences.
The 25 questions cover a diverse range of ages, genders, and ethnicities, and the candidate images in each question include both clearly different cases and more subtle differences in texture and details.
To reduce position bias, the order of the five candidate images within each question is randomized. The user interface of Questionnaire~I is illustrated in Figure~\ref{supp:survey_1_ui}.

\begin{figure}[tbp] 
\begin{center}
\includegraphics[width=\linewidth]{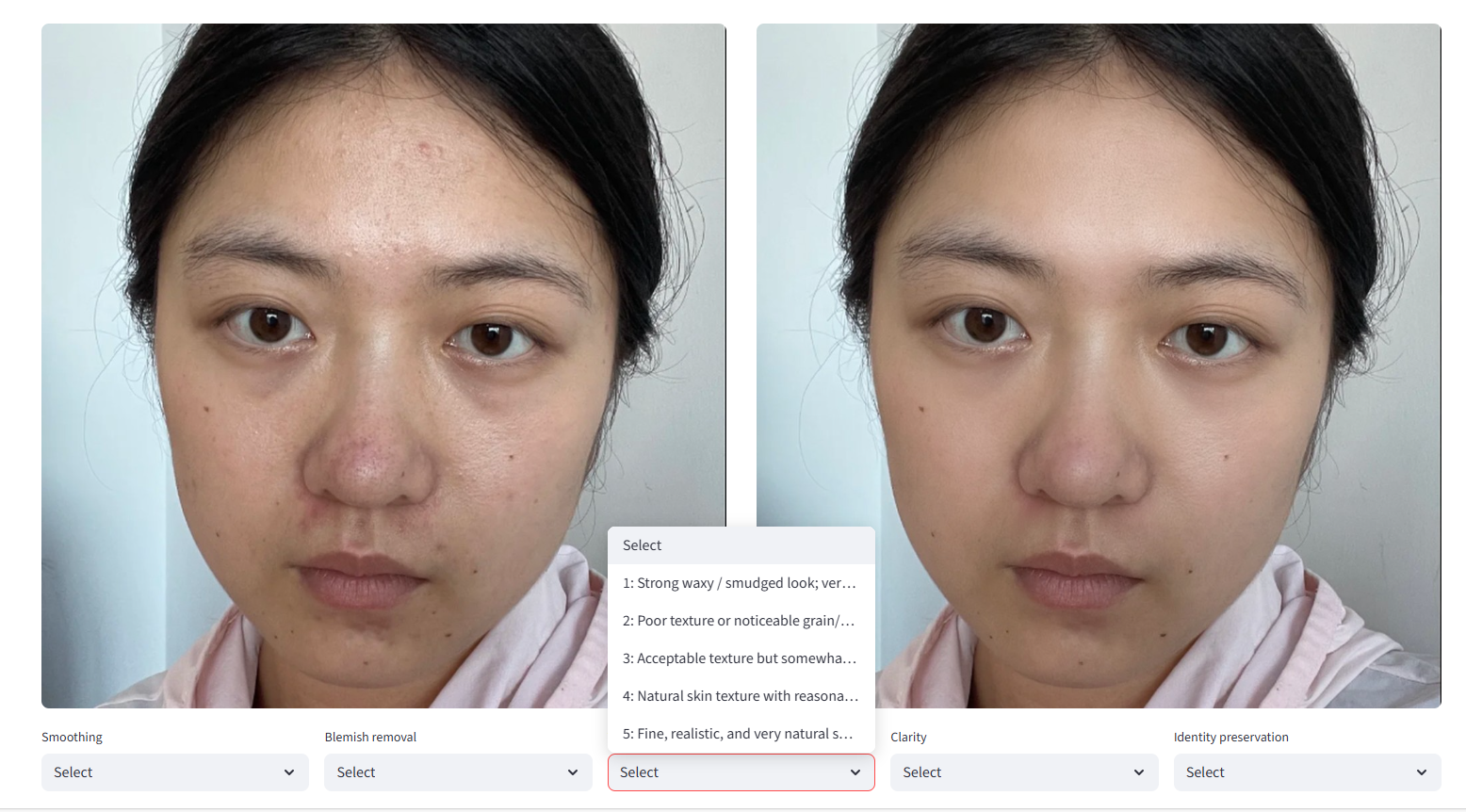}
\end{center}
\caption{User interface of Questionnaire~II.}
\label{supp:survey_2_ui}
\end{figure}

\textbf{Questionnaire II: Dimension-wise Ratings for Reward Evaluation.}
The second questionnaire is designed to measure the consistency between different reward models and human judgments on fine-grained perceptual dimensions.
It is constructed from an additional pool of 20 input–output pairs. For each participant, 5 questions are randomly sampled from this pool.
Each question shows one original portrait and one retouched result, and users rate the result along five dimensions:
skin smoothing, blemish removal, skin texture quality, clarity, and identity preservation.
For each dimension, participants assign an integer score in the range $0$–$5$, where a larger value indicates better perceived quality.
We provide a detailed textual description and concrete examples for every score level of each dimension (from $0$ to $5$), so that users have a clear and consistent guideline when rating.
These human scores are later compared with the outputs of different reward models to quantify their alignment with human perception. The user interface of Questionnaire~II is illustrated in Figure~\ref{supp:survey_2_ui}.

\textbf{Data Collection and Aggregation.}
In total, we obtain 100 valid questionnaires from anonymous participants.
For Questionnaire~I, we compute the win rate of each method as the proportion of answers in which it is selected as the preferred result over all questions and participants.
For Questionnaire~II, we aggregate the human scores for each dimension by averaging over all participants and questions, and then compare these averages with the corresponding scores predicted by different reward models.
This allows us to evaluate dimension-wise consistency between reward models and human judgments.

\begin{figure}[tbp] 
\begin{center}
\includegraphics[width=\linewidth]{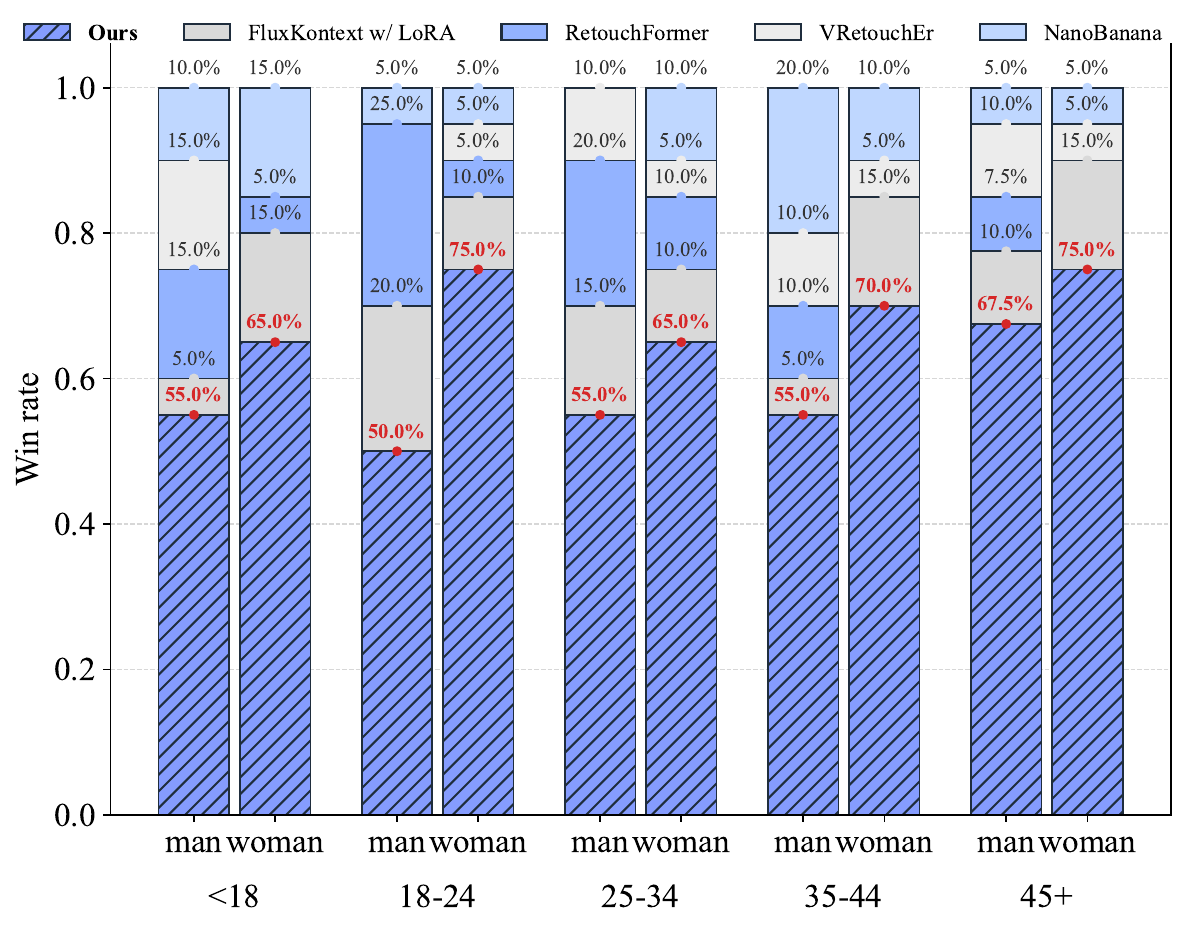}
\end{center}
\caption{Comparison of user preference win rates for various face retouching methods across different ages and gender.}
\label{supp:survey_1}
\end{figure}

\subsection{Demographic Analysis of User Preferences}
Figure~\ref{supp:survey_1} presents the win rates for Questionnaire~I, stratified by user age group and gender. Our method achieves the highest win rate in every demographic group, consistently capturing at least 50\% of user preferences and reaching about 75\% for women aged 18–24 and 45+ years. Competing methods obtain much smaller shares, with none exceeding 25\% in any group or showing consistent dominance for any demographic. These results indicate that the preference advantage of our method is robust across ages and genders, rather than being driven by a single specific subgroup.

\section{Training Details}
\subsection{Reward Model Training Details}
We adopt Qwen2.5-VL-7B-Instruct \cite{bai2025qwen25vl} as our base reward model. For both SFT stages (Stage~1 and Stage~2), training is performed with a per-device batch size of $1$, $8$ gradient accumulation steps, a learning rate of $5\times10^{-5}$, and a warm-up ratio of $0.1$.
For the GRPO stage, we use a per-device batch size of $2$, $4$ gradient accumulation steps, a learning rate of $1\times10^{-6}$, and a warm-up ratio of $0.05$, with a KL penalty coefficient $\beta = 0.04$. The number of generated responses $N$ is set to $8$.

\subsection{Retouching Model Training Configuration}
We adopt Flux.1~Kontext \cite{fluxkontext2025} as our base retouching model and first perform LoRA adaptation on our face retouching corpus. The model is trained at a resolution of $1024\times1024$ with LoRA rank $32$ and $\alpha=32$, using AdamW with a learning rate of $1\times10^{-5}$, cosine learning rate decay, $500$ warm-up steps, weight decay $0.01$, a per-device batch size of $4$, and $2$ gradient accumulation steps.
For BeautyGRPO, we initialize from this LoRA-adapted Flux.1~Kontext and run online RL at $1024\times1024$ resolution with a per-device batch size of $2$ and $8$ images per prompt, using $T=10$ sampling steps and a noise level $\eta=0.7$.

\section{CoT Reasoning Examples of Reward Model}
\begin{figure*}[htbp] 
\begin{center}
\includegraphics[width=\linewidth]{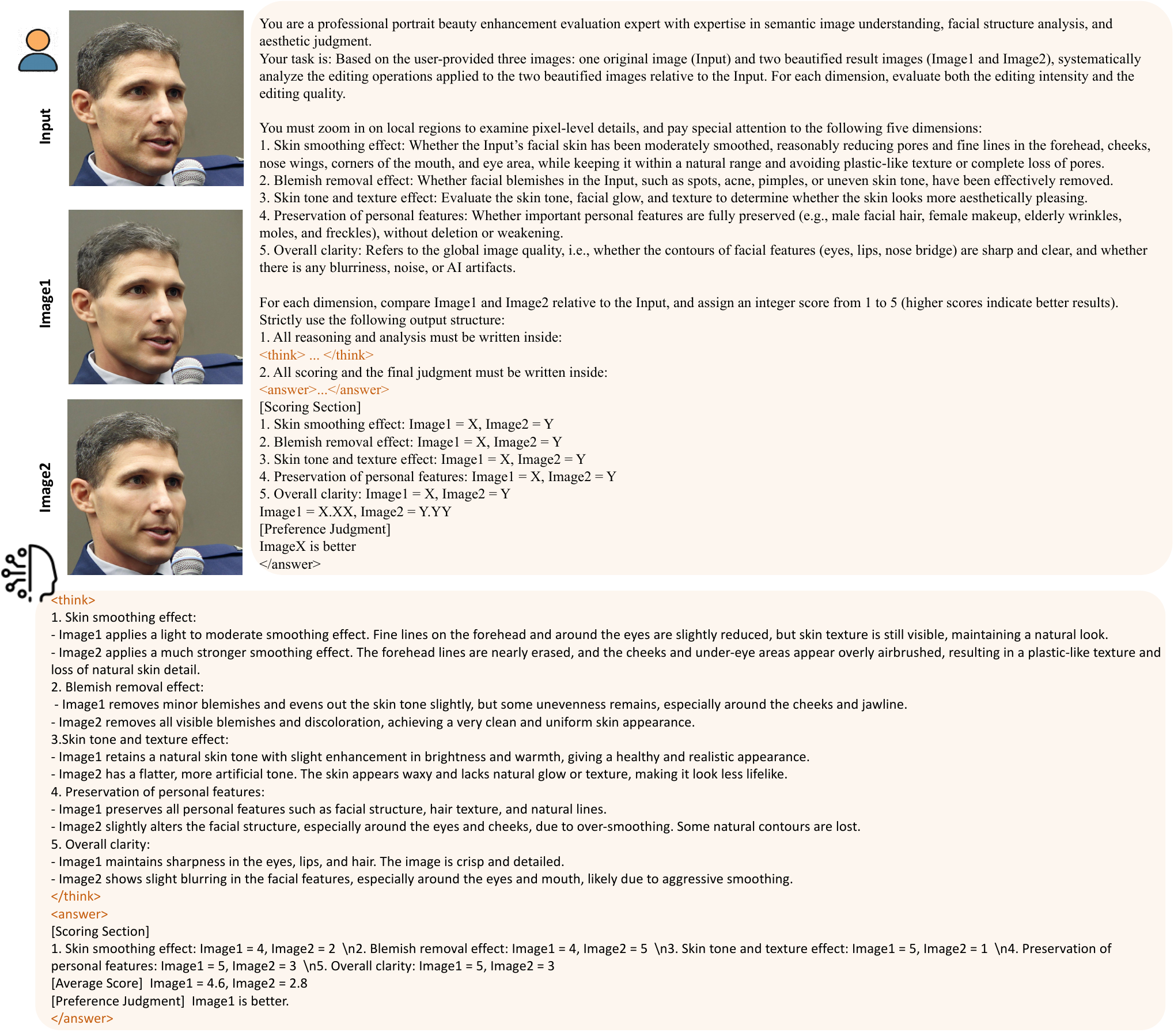}
\end{center}
\caption{Qualitative examples of face retouching CoT reward reasoning. Given an input image and two retouched candidates, our reward model performs chain-of-thought quality assessment across five dimensions: skin smoothing, blemish removal, skin texture quality, clarity, and identity preservation.}
\label{supp:template}
\end{figure*}

We provide qualitative examples of our reward model's Chain-of-Thought (CoT) reasoning process in Figure~\ref{supp:template}.

\section{More Visual Results}
\begin{figure*}[htbp] 
\begin{center}
\includegraphics[width=\linewidth]{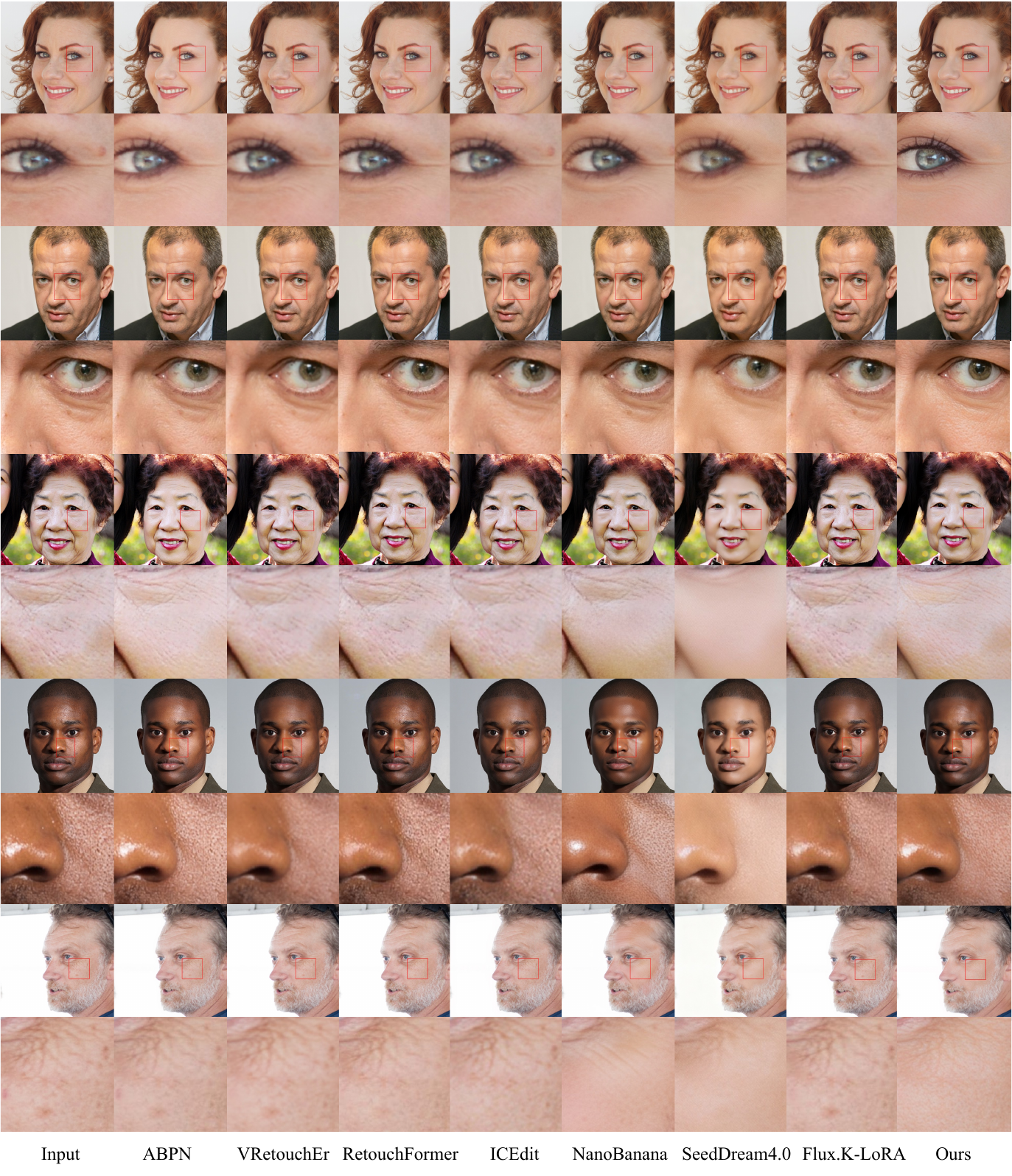}
\end{center}
\caption{Visual comparison of face retouching results across different methods on the FFHQR dataset, where Flux.K denotes FluxKontext. 
Compared with existing methods that suffer from incomplete blemish removal, over-smoothing, or unnatural visual appearance, our BeautyGRPO cleanly removes blemishes while preserving identity-consistent natural skin texture, gloss, and fine details such as wrinkles.}
\label{supp:res1}
\end{figure*}

\begin{figure*}[htbp] 
\begin{center}
\includegraphics[width=\linewidth]{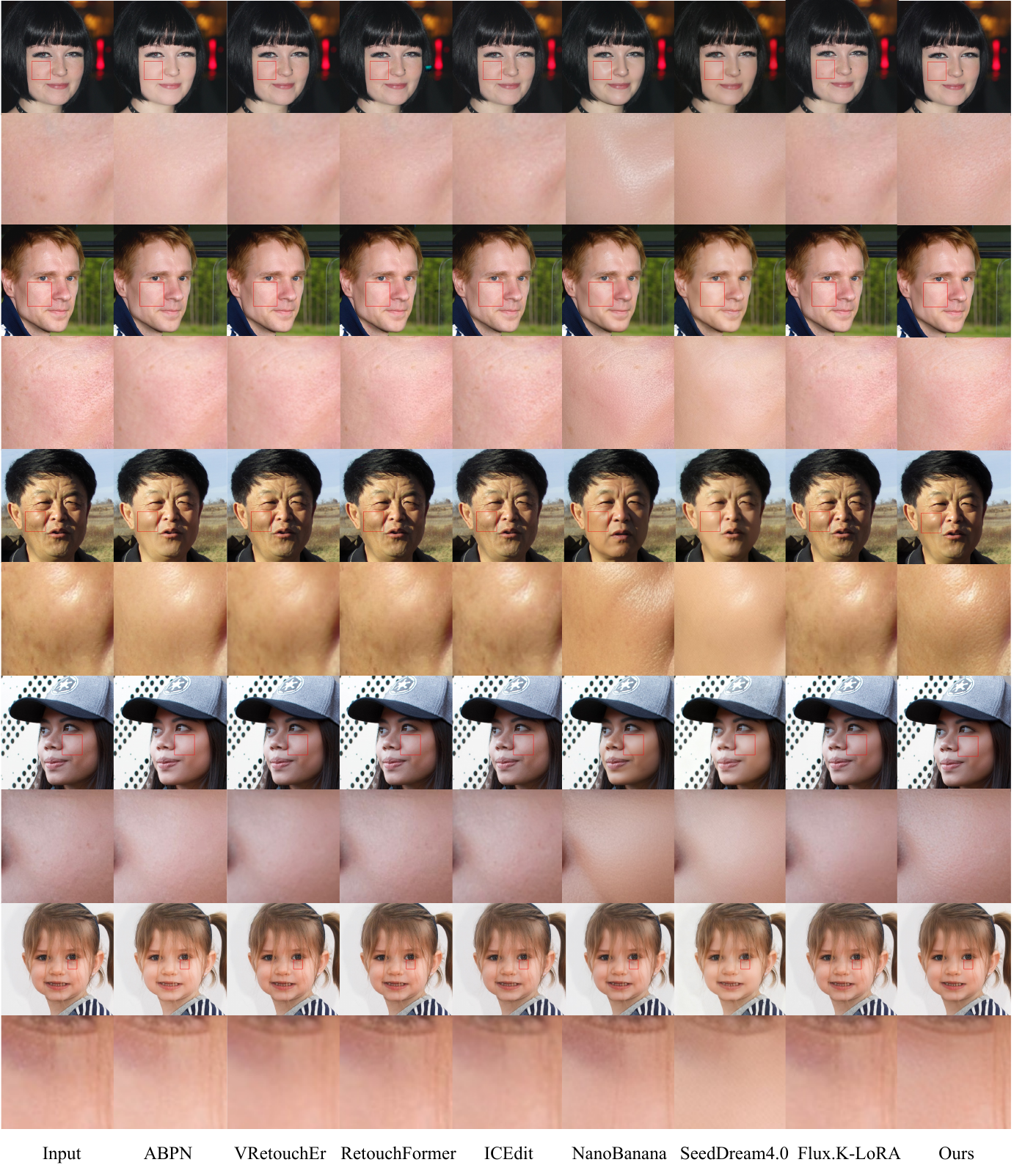}
\end{center}
\caption{Visual comparison of face retouching results across different methods on the FFHQR dataset, where Flux.K denotes FluxKontext. 
Compared with existing methods that suffer from incomplete blemish removal, over-smoothing, or unnatural visual appearance, our BeautyGRPO cleanly removes blemishes while preserving identity-consistent natural skin texture, gloss, and fine details such as wrinkles.}
\label{supp:res2}
\end{figure*}

\begin{figure*}[htbp] 
\begin{center}
\includegraphics[width=\linewidth]{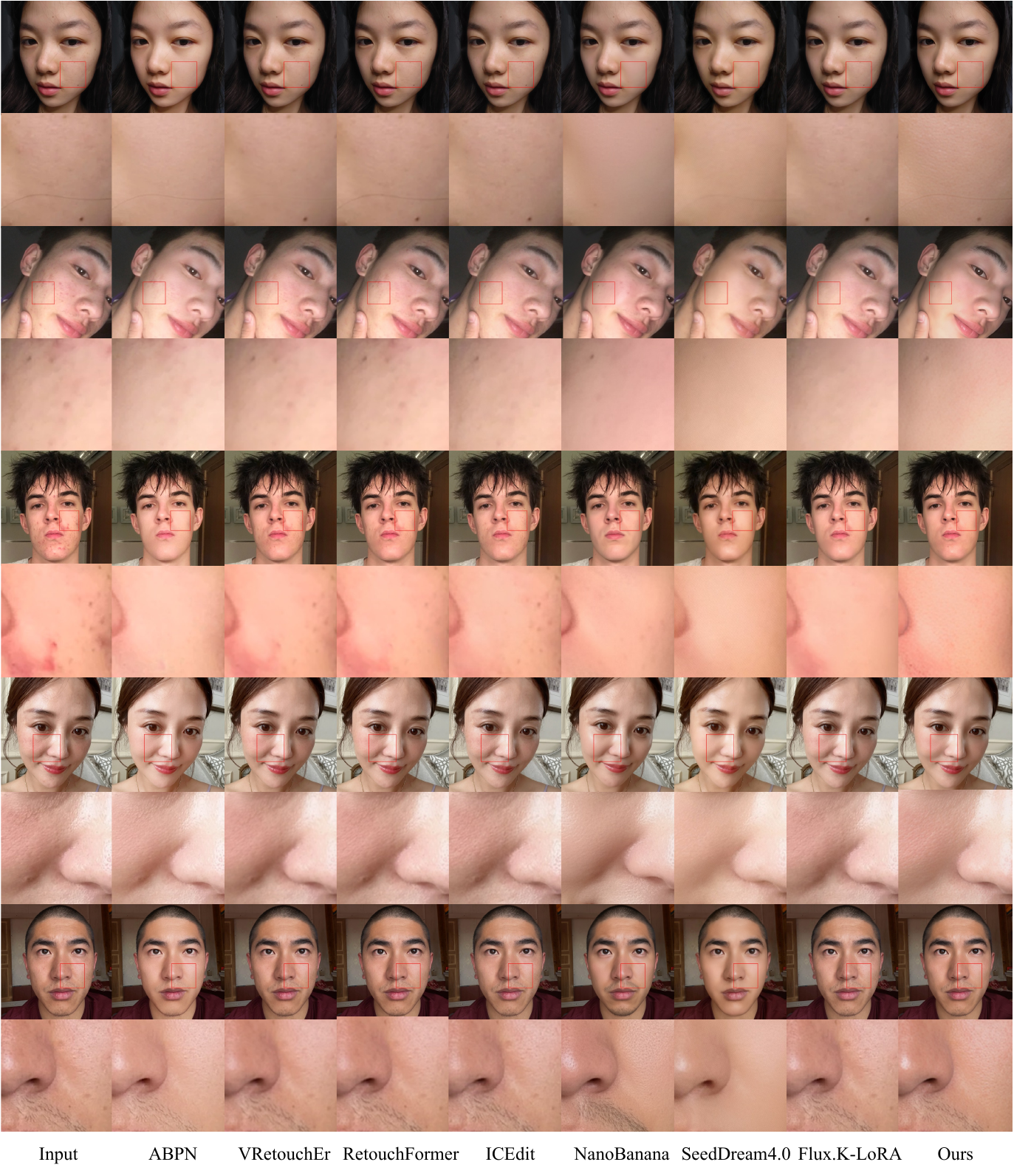}
\end{center}
\caption{Visual comparison of face retouching results across different methods on in-the-wild  datasets, where Flux.K denotes FluxKontext. 
Compared with existing methods that suffer from incomplete blemish removal, over-smoothing, or unnatural visual appearance, our BeautyGRPO cleanly removes blemishes while preserving identity-consistent natural skin texture, gloss, and fine details such as moles.}
\label{supp:res3}
\end{figure*}

Figures~\ref{supp:res1}--\ref{supp:res3} provide additional qualitative comparisons across different face retouching methods, further highlighting the perceptual advantages of our BeautyGRPO.

\section{Discussion and Future Work}
\label{sec:discussion_limitations}

\subsection{Anchor Selection and Quality}

In our main experiments, we use the top-1 high-preference sample as the anchor $x_0^{\text{anchor}}$ for each input, which keeps the implementation simple and ensures that the anchor lies on a high-fidelity manifold. Conceptually, DPG only requires the anchor to be a reasonably good exemplar that constrains exploration to a high-quality neighborhood while still allowing the policy to move beyond it when the reward signal suggests a better solution. Systematic studies of alternative anchor selection strategies (e.g., sampling from a top-$k$ set, diversity-aware anchors, or combining multiple anchors) are promising directions for future work.

\subsection{Scenarios Without Explicit Anchors}

In this work, we focus on a practical face-retouching regime where each input portrait is accompanied by at least one high-quality retouched exemplar (e.g., curated or highly rated images in FRPref-10K), which naturally provides anchors for DPG. In scenarios where only raw inputs are available and no such exemplars exist, one could instead construct pseudo-anchors using a strong reward model, self-distilled EMA policies, or other generative priors, and then apply the same anchored-exploration principle. Exploring such anchor-free or weakly anchored settings is an interesting extension of DPG that we leave to future work.

\subsection{DPG at Inference Time}

We apply DPG only during the sampling-based exploration phase of RL training. After optimization, standard ODE sampling uses the improved vector field to produce high-fidelity results without anchors or extra computation at inference time. In principle, one could design inference-time variants of DPG by supplying reference images or synthesizing anchors on the fly, but this would introduce additional assumptions and complexity. We therefore adopt pure ODE sampling for deployment and leave inference-time extensions of DPG as future work.

\end{document}